\documentclass[smallcondensed]{svjour3}

\usepackage{amssymb}
\usepackage{amsmath}
\usepackage{cases}
\usepackage{graphicx}
\usepackage{epstopdf}
\usepackage{hyperref}
\usepackage{array}
\usepackage{natbib}
\usepackage{tab}

\renewcommand{\epsilon}{\varepsilon}

\newcommand{\cA}{\mathcal{A}}
\newcommand{\cT}{\mathcal{T}}

\newcommand{\cL}{\mathcal{L}}
\newcommand{\cC}{\mathcal{C}}

\newcommand{\cE}{\mathcal{E}}

\newcommand{\cS}{\mathcal{S}}
\newcommand{\cF}{\mathcal{F}}
\newcommand{\cG}{\mathcal{G}}
\newcommand{\cW}{\mathcal{W}}
\newcommand{\cX}{\mathcal{X}}
\newcommand{\cH}{\mathcal{H}}
\newcommand{\cU}{\mathcal{U}}

\newcommand{\abs}[1][\,\cdot\,]{\ensuremath{| #1 |}}
\newcommand{\bp}{\boldsymbol{p}}
\newcommand{\bq}{\boldsymbol{q}}
\newcommand{\bw}{\boldsymbol{w}}

\newcommand{\R}{\mathbb{R}}
\renewcommand{\leq}{\leqslant}
\renewcommand{\geq}{\geqslant}
\newcommand{\wh}{\widehat}
\newcommand{\wt}{\widetilde}
\newcommand{\rmse}{\mbox{\sc rmse}}

\newcommand{\size}{\mbox{size}}
\newcommand{\ind}{\mathbb{I}}
\newcommand{\grad}{\mbox{\rm \tiny grad}}
\newcommand{\argmin}{\mathop{\mathrm{argmin}}}
\newcommand{\tl}{\widetilde{\ell}}
\newcommand{\tR}{\widetilde{R}}
\newcommand{\Tr}{\mathrm{Tr}}

\newlength{\minipagewidth}
\newcommand{\bookbox}[1]{
\par\medskip\noindent
\framebox[\textwidth]{
\begin{minipage}{\minipagewidth} #1 \end{minipage} } \par\medskip
}

\begin{document}

\title{Forecasting electricity consumption by aggregating specialized experts}
\subtitle{A review of the sequential aggregation of specialized experts, with an
application to Slovakian and French country-wide one-day-ahead (half-)hourly predictions}

\author{Marie Devaine \and Pierre Gaillard \and Yannig Goude \and Gilles Stoltz}

\institute{M. Devaine \at
Ecole Normale Sup{\'e}rieure, Paris, France \\
\email{marie.devaine@ens.fr}
\and
P. Gaillard \at
Ecole Normale Sup{\'e}rieure, CNRS, INRIA, Paris, France \\
Tel.: +33-144-322-041 \\
\email{pierre.gaillard@ens.fr}
\and
Y. Goude \at
EDF R{\&}D, Clamart, France \\
Tel.: +33-147-651-561 \\
\email{yannig.goude@edf.fr}
\and
G. Stoltz \at
Ecole Normale Sup{\'e}rieure, CNRS, INRIA, Paris, France \\
\& \\
HEC Paris, CNRS, Jouy-en-Josas, France \\
Tel.: +33-144-323-277 \\
\email{gilles.stoltz@ens.fr}
}

\date{Received: 27 March 2011 / Revised: 10 February 2012 / Accepted: 4 June 2012}

\maketitle

\begin{abstract}
We consider the setting of sequential prediction of arbitrary sequences based on specialized experts.
We first provide a review of the relevant literature and present two theoretical contributions:
a general analysis of the specialist aggregation rule of \cite{FrScSiWa97} and
an adaptation of fixed-share rules of~\cite{HeWa98} in this setting.
We then apply these rules to the sequential short-term (one-day-ahead) forecasting of electricity consumption;
to do so, we consider two data sets, a Slovakian one and a French one, respectively concerned with hourly and half-hourly predictions.
We follow a general methodology to perform the stated empirical studies
and detail in particular tuning issues of the learning parameters.
The introduced aggregation rules demonstrate an improved accuracy on the data sets at hand;
the improvements lie in a reduced mean squared error but also in a more robust behavior with respect to large occasional errors.
\end{abstract}

\section{Introduction and motivation}

We consider the sequential prediction of arbitrary sequences based on expert advice,
the topic of a large literature summarized in the monography of~\cite{CeLu06}.
At each round of a repeated game of prediction, experts output forecasts, which are to be
combined by an aggregation rule (usually based on their past performance); the true outcome
is then revealed and losses, which correspond to prediction errors, are suffered by
the aggregation rules and the experts. We are interested in aggregation rules that perform
almost as well as, for instance, the best constant convex combination of the experts.
In our setting, these guarantees are not linked in any sense to a stochastic model: in fact, they hold
for all sequences of consumptions, in a worst-case sense.

The application we have in mind --the sequential short-term (one-day-ahead)
forecasting of electricity consumption--
will take place in a variant of the basic problem of prediction
with expert advice called prediction with specialized (or sleeping) experts.
At each round only some of the experts output a prediction while the other ones are inactive.
This more difficult scenario does not arise from experts being lazy but rather from them
being specialized. Indeed, each expert is expected to provide accurate forecasts mostly in given
external conditions, that can be known beforehand.
For instance, in the case of the prediction of electricity consumption, experts
can be specialized to winter or to summer, to working days or to public holidays, etc.

The literature on specialized experts is --to the best of our knowledge-- rather sparse.
The first references are~\cite{Bl97} and~\cite{FrScSiWa97};
they respectively introduce and formalize the framework of specialized experts.
They were followed only by few other ones: two papers mention some results
for the context of specialized experts only in passing (\cite[Sections~6--8]{BlMa07}
and \cite[Section~6.2]{CeLu03}) while another one considers a somewhat different
notion of regret, namely, \cite{KlNiSh08}.

The theory of prediction with expert advice has of course been already
applied to real data in many fields; we provide a list and a classification
of such empirical studies in Section~\ref{sec:calibr}.
We only mention here that as far as the forecasting of electricity consumption is concerned,
a preliminary study of some aggregation rules for individual sequences
was already performed for the daily prediction of the French electricity
load in~\cite{Gou08,Gou08_2}.

\subsubsection*{Contributions and outline of the paper}
We review in Section~\ref{sec:theory} the framework of
sequential prediction with specialized experts.
Three families of aggregation rules are discussed,
which were for two of them obtained by
taking a new look at existing strategies; this new look corresponds
to (slight or more important) adaptations of these existing strategies
and to simpler or more general analyses of their theoretical performance bounds.
Finally, a practical online tuning of these aggregation rules is developed and put
in perspective with respect to theoretical methods to do so.

We then study, respectively in Sections~\ref{sec:slov} and~\ref{sec:French},
the performance obtained by the developed aggregation rules on two data sets. The first one
was provided by the Slovakian subbranch of EDF (``Electricit{\'e} de France'',
a French electricity provider) and represents its local market; the second one deals with the French market
for which EDF is still the overwhelming provider. These empirical studies
are organized according to the same standardized methodology described in Section~\ref{sec:methodo}:
construction of the experts based on historical data;
tabulation of the performance of some benchmark prediction methods;
results obtained by the sequential aggregation rules, first with parameters optimally tuned in hindsight,
and then when the tuning is performed sequentially according to the introduced online tuning.
The section on French data is also followed by a note (Section~\ref{sec:robust})
on the individual performance of the aggregation rules, i.e., an indication that
their behavior is not only good on average but also that the large prediction errors
occur less frequently for the aggregation rules than for the base experts.

\section{Aggregation of specialized experts: A survey with some new results}
\label{sec:theory}
\label{sec:deffam}

The following framework was introduced in~\cite{Bl97} and further studied in~\cite{FrScSiWa97}.

A bounded sequence of observations (e.g., hourly or half-hourly electricity consumptions)
$y_1,y_2,\ldots,y_T \in [0,B]$ is to be predicted element by element at time instances $t = 1,2,\ldots,T$.
A finite number $N$
of base forecasting methods, henceforth referred to as experts, are
available. Before each time instance $t$, some experts provide a forecast and the other ones do not.
The first ones are said active and their forecasts are denoted by
$f_{j,t} \in \R_+$, where $j$ is the index of the considered active expert;
the experts of the second group are said inactive.
We assume that the experts know the bound $B$ and only produce
forecasts $f_{j,t} \in [0,B]$.
Finally, we denote by $E_t \subset \{ 1, \ldots, N \}$ the set of active experts
at a given time instance $t$ and assume that it is always non empty.

At each time instance $t \geq 1$, a sequential convex aggregation rule produces a
convex weight vector $\bp_t = (p_{1,t},\ldots,p_{N,t})$ based on the past
observations $y_1,\ldots,y_{t-1}$ and the past
and present forecasts $f_{j,s}$, for all $s = 1,\ldots,t$ and $j \in E_s$.
By convex weight vector, we mean a vector $\bp_t \in \R^N$ such that
$p_{j,t} \geq 0$ for all $j = 1,\ldots,N$ and $p_{1,t} + \ldots + p_{N,t} = 1$;
we denote by $\cX$ the set of all these convex weight vectors over $N$ elements.
The final prediction at $t$ is then obtained by linearly combining the
predictions of the experts in $E_t$ according to the weights given by the components of
the vector $\bp_t$. More precisely, the aggregated prediction at time instance $t$ equals
\[
\widehat{y}_t = \sum_{j \in E_t} p_{j,t} f_{j,t}\,.
\]
The observation $y_t$ is then revealed and instance $t+1$ starts.

To measure the accuracy of the prediction $\wh{y}_t$ proposed at round $t$
for the observation $y_t$ we consider a loss function $\ell : \R \times \R
\to \R$. At each time instance $t$, the convex combination $\bp_t$ output by the rule
is thus evaluated by the loss function $\ell_t : \cX \to \R$ defined by
\[
\ell_t(\bp) = \ell\!\left(\sum_{j \in E_t} p_j f_{j,t}, \, y_t \right)
\]
for all $\bp \in \cX$. The subscript $t$ in the notation $\ell_t$ encompasses
the dependencies in the expert forecasts $f_{j,t}$ and in the outcome $y_t$.
Our goal is to design sequential convex aggregation rules $\cA$ with a small cumulative
error $\sum_{t=1}^T \ell_t(\bp_t)$.
To do so, we will ensure that quantities called regrets (with respect to fixed
experts, to fixed convex combinations of experts, or to sequences of experts with few shifts)
are small.

Possible loss functions are the square loss, defined by $\ell(x,y) = (x-y)^2$
for all $x,y \in [0,B]$, the absolute loss $\ell(x,y) = |x-y|$,
and the absolute percentage of error $\ell(x,y) = |x-y|/y$, which are all three convex
and bounded (so that their associated loss functions $\ell_t$ are convex and bounded as well).

\subsection{Minimizing regret with respect to fixed experts}
\label{sec:defEWA}

This notion of regret was introduced in~\cite{FrScSiWa97}
and compares the error suffered by a rule $\cA$
to the one of a given expert $j$ only on time instances when $j$ was active;
formally, the regret of $\cA$ with respect to expert $j$ up to instance $T$ equals
\begin{equation}
\label{eq:defregr1}
R_T(\cA,j) = \sum_{t=1}^T \bigl( \ell_t(\bp_t) - \ell_t(\delta_j) \bigr) \, \ind_{ \{ j \in E_t \} }\,,
\end{equation}
where $\delta_j \in \cX$ is the Dirac mass on $j$ (the convex weight vector with weight $1$ on $j$).

\subsubsection*{The exponentially weighted average aggregation rule}
It relies on a parameter $\eta > 0$ and will thus be denoted by
$\cE_\eta$. It chooses $\bp_1$ to be the uniform distribution over $E_1$
and uses at time instance $t \geq 2$ the convex weight vector $\bp_t$
given by
\begin{equation}
\label{eq:defEWA}
p_{j,t} = \frac{e^{\eta R_{t-1}( \cE_\eta, j)} \, \ind_{ \{ j \in E_t \} }}{\sum_{k \in E_t}
e^{\eta R_{t-1}( \cE_\eta, k)}}\,;
\end{equation}
that is, it only puts mass on the experts $j$ active at round $t$ and does so by
performing an exponentially weighted average of their past performance, measured by the regrets
$R_{t-1}( \cE_\eta, j)$.

The following performance bound
is a straightforward consequence of the results presented in~\cite{CeLu03} (its Corollary~2
and the methodology followed in its Sections~3 and~6.2).

\begin{theorem}
\label{prop:ewa}
We assume that the loss functions $\ell_t$ are convex and uniformly bounded; we denote by $L$ a uniform
bound on the quantities $|\ell_t(\delta_i) - \ell_t(\delta_j)|$ when $i$ and $j$ vary in $E_t$ and $t$ varies
from $1$ to $T$. The regret of $\cE_\eta$ is bounded over
all such sequences of expert forecasts and observations as
\begin{equation}
\label{eq:regretboundEWA}
\max_{j = 1,\ldots,N} R_{T}( \cE_\eta, j) \leq \frac{\ln N}{\eta} + \frac{\eta}{2} L^2 T\,.
\end{equation}
\end{theorem}

The (theoretically) optimal choice $\eta^\star = \sqrt{(2 \ln N)/(L^2 T)}$
leads to the uniform bound $L \sqrt{2 T \ln N}$ on the regret of $\cE_{\eta^\star}$.
This choice depends on the horizon $T$ and of the bound $L$, which are not always known in advance;
standard techniques, like the doubling trick or time-varying learning rates $\eta_t$ can be used
to cope with these limitations as far as theoretical bounds are concerned, see~\cite{AuCeGe02,CeMaSt07}.

\begin{remark}
A slightly different family of aggregation rules based on exponentially weighted averages, referred to as
$\cH$ in the sequel (which stands for Hedge), was presented in~\cite[Section~6]{BlMa07}.
It replaces the update~\eqref{eq:defEWA} by
\begin{multline}
\nonumber
w_{j,t} = \exp \! \left( -\eta_j \sum_{s=1}^{t-1} \bigl( \ell_s(\delta_j) - e^{-\eta_j} \ell_s(\bp_s) \bigr) \right) \\
\mbox{and} \qquad
p_{j,t} = \frac{w_{j,t} \bigl( 1-e^{-\eta_j} \bigr) \, \ind_{ \{ j \in E_t \} }}{
\sum_{k \in E_t} w_{k,t} \bigl( 1-e^{-\eta_k} \bigr) \, \ind_{ \{ k \in E_t \} }}\,,
\end{multline}
where the learning rates $\eta_j$ now depend on the experts $j = 1,\ldots,N$.
By carefully setting these rates, uniform regret bounds of the form
\[
R_{T}( \cH, j) = \mathcal{O} \! \left( L \, \sqrt{\sum_{t=1}^T \ind_{ \{ j \in E_t \} } \ln N} + L \ln N \right)
\]
can be obtained.
However, we checked in \cite[Section~2.1]{DeGoSt09} that the empirical performance of the families of rules $\cH$
and $\cE$ were equal. This is why only the simplest of the two, $\cE$, will be considered in the sequel.
\end{remark}

\subsubsection*{The specialist aggregation rule}

The content of this section revisits and (together with the gradient trick recalled in
the next section) improves on the results of~\cite[Sections~3.2--3.4]{FrScSiWa97}.
In the latter reference, aggregation rules designed to minimize the regret were
introduced but their statement, analyses, and regret bounds heavily depended on the specific\footnote{See equation (6) in \cite{FrScSiWa97}
and the comments after its statement: ``Here, $a$ and $b$ are positive constants which depend on the
specific on-line learning problem [...].''} loss functions at
hand. Two special cases were worked out (absolute loss and square loss).
In contrast, we provide a compact and general analysis, solely based on Hoeffding's lemma.

The specialist aggregation rule is described in Figure~\ref{fig:Srule}; it
relies on a parameter $\eta > 0$ and will be denoted by $\cS_\eta$.
It is close in spirit to but different from the rule $\cE_\eta$: as we will see below,
the two rules have comparable theoretical guarantees, their statements might be found to exhibit some similarity as well,
but we noted that in practice the output convex weight vectors $\bp_t$ had little in common (even though
the achieved performance was often similar).
\begin{figure}[t]
\emph{Parameters}: learning rate $\eta > 0$
\smallskip\noindent

\emph{Initialization}: $\bw_1$ is the uniform convex weight vector, $w_{i,1} = 1/N$ for $i=1,\ldots,N$
\smallskip\noindent

{\em For} each time instance $t = 1,2,\ldots,T$,
\smallskip\noindent
\begin{itemize}
  \item[(1)] predict $\displaystyle{\wh{y}_t = \frac{1}{\sum_{k \in E_t} w_{k,t}} \, \sum_{j \in E_t} w_{j,t} \, f_{j,t}}$\,;
  \item[(2)] observe $y_t$ and compute the convex weight vector $\bw_{t+1}$ as
  \begin{numcases}{w_{i,t+1} = }
  \nonumber
  w_{i,t} \, e^{-\eta \ell_t(\delta_i)}
  \frac{\sum_{j\in E_t} w_{j,t}}{\sum_{k\in E_t} w_{k,t} \, e^{-\eta \ell_t(\delta_k)}} & if $i\in E_t$, \\
  \nonumber
  w_{i,t} & if $i \not\in E_t$.
  \end{numcases}
  \end{itemize}
\rule{\linewidth}{.5pt}
\caption{\label{fig:Srule} The specialist aggregation rule $\cS_\eta$.}
\end{figure}

\begin{theorem}
\label{th:spec}
We assume that the loss functions $\ell_t$ are convex and uniformly bounded; we denote by $L$ a constant
such that the quantities $\ell_t(\delta_i)$ all belong to $[0,L]$ when $i$ varies in $E_t$ and $t$ varies
from $1$ to $T$. The regret of $\cS_\eta$ is bounded over
all such sequences of expert forecasts and observations as
\[
\max_{j = 1,\ldots,N} R_{T}( \cS_\eta, j) \leq \frac{\ln N}{\eta} + \frac{\eta}{8} L^2 T\,.
\]
\end{theorem}

The proof of this theorem is postponed to the appendix (Section~\ref{sec:app2}).
The (theoretically) optimal choice $\eta^\star = \sqrt{(8 \ln N)/(L^2 T)}$
leads to the uniform bound $L \sqrt{(T/2) \ln N}$ on the regret of $\cS_{\eta^\star}$.
The same comments on the calibration of $\eta$ as in the previous sections apply.

\subsection{Minimizing regret with respect to fixed convex combinations of experts}
\label{sec:subgrad}

This notion of regret was introduced in~\cite{FrScSiWa97} as well
and compares the error suffered by a rule $\cA$
to the one of a given convex combination $\bq \in \cX$ as follows.
Formally, for a set $E \subset \{ 1,\ldots,N \}$ of active experts,
we define
\[
\bq(E) = \sum_{j \in E} q_j
\]
and denote by $\bq^E = (q_1^E, \ldots, q_N^E)$ the convex weight vector obtained by
``conditioning'' $\bq$ to $E$:
\begin{numcases}{\bq^E =}
\nonumber
(0,\ldots,0) & if $\bq(E) = 0$; \vspace{.15cm} \\
\nonumber
\left( \frac{q_1 \ind_{ \{ 1 \in E \} }}{\bq(E)}, \, \ldots, \,
\frac{q_N \ind_{ \{ N \in E \} }}{\bq(E)} \right) & if $\bq(E) > 0$.
\end{numcases}
Now, the definition~(\ref{eq:defregr1}) can be generalized as
\begin{equation}
\label{eq:defregr2}
R_T(\cA,\bq) = \sum_{t=1}^T \Bigl( \ell_t(\bp_t) - \ell_t\bigl(\bq^{E_t}\bigr) \Bigr) \, \bq(E_t)\,.
\end{equation}
This is indeed a generalization as we have $R_T(\cA,\delta_j) = R_T(\cA,j)$.

We deal with this more ambitious goal by resorting to the so-called gradient trick,
see~\cite[Section 2.5]{CeLu06} for more details.
When the loss function $\ell : [0,B]^2 \to \R$
is convex and (sub)differentiable in its first argument,
then the functions $\ell_t$ are convex and (sub)differentiable over $\cX$;
we denote by $\nabla\ell_t$ their (sub)gradient function. By denoting by $\cdot$
the inner product in $\R^N$ and viewing $\cX$ as a subset of $\R^N$, we have the following inequality:
for all $t$, for all $\bq \in \cX$,
\[
\ell_t(\bp_t) - \ell_t(\bq) \leq \nabla\ell_t(\bp_t) \,\cdot\, \bigl( \bp_t - \bq \bigr)
= \tl_t(\bp_t) - \tl_t(\bq)\,,
\]
where we denoted by $\tl_t(\bq) = \nabla\ell_t(\bp_t) \,\cdot\, \bq$
the pseudo-loss function associated with time instance $t$. It is linear over $\cX$.
Now, the gradient trick simply consists of replacing the loss functions $\ell_t$
by the pseudo-loss functions $\tl_t$ in the definitions of the forecasters.
In particular, this replacement in~\eqref{eq:defEWA}, where the loss functions are hidden in the regret terms,
respectively, in Figure~\ref{fig:Srule},
leads to an aggregation rule denoted by $\cE_\eta^{\grad}$,
respectively, $\cS_\eta^{\grad}$.

Now, the above convexity inequality and the linearity of the $\tl_t$
imply that for any rule $\cA$,
\begin{multline}
\nonumber
\max_{\bq \in \cX} R_{T} \bigl( \cA, \bq \bigr)
\leq \max_{\bq \in \cX} \, \sum_{t=1}^T \Bigl( \tl_t(\bp_t) - \tl_t\bigl(\bq^{E_t}\bigr) \Bigr) \, \bq(E_t) \\
= \max_{\bq \in \cX} \, \sum_{j=1}^N q_j \, \sum_{t=1}^T \Bigl( \tl_t(\bp_t) - \tl_t(\delta_j) \Bigr) \,
\ind_{ \{ j \in E_t \} }\,;
\end{multline}
the following result is thus a corollary of Theorems~\ref{prop:ewa} and~\ref{th:spec}.

\begin{corollary}
\label{cor:ewa-spec}
We assume that the loss functions $\ell_t$ are convex and (sub)differentiable over $\cX$,
with (sub)gradient functions uniformly bounded in the supremum norm as $t$ varies by $G$.
The regret of $\cE^{\grad}_\eta$ is bounded over
all such sequences of expert forecasts and observations as
\[
\max_{\bq \in \cX} R_{T} \bigl( \cE^{\grad}_\eta, \bq \bigr) \leq \frac{\ln N}{\eta} + 2 \eta G^2 T
\]
while the one of $\cS^{\grad}_\eta$ is also uniformly bounded as
\[
\max_{\bq \in \cX} R_{T} \bigl( \cS^{\grad}_\eta, \bq \bigr) \leq \frac{\ln N}{\eta} + \frac{\eta}{2} G^2 T\,.
\]
\end{corollary}

\subsection{Minimizing regret with respect to sequences of (convex combinations of) experts with few shifts}
\label{sec:defFS}

This third and last definition of regret was introduced by~\cite{HeWa98} and
compares the performance of a rule not to the performance of a fixed expert or a fixed convex combination
of the experts, but to sequences of experts or of convex combinations of experts (abiding by the activeness
constraints given by the $E_t$).
To the best of our knowledge, this approach of considering sequences of experts
had not been used before to deal with specialized experts.

Formally, we denote by $\cL$ the set of all legal sequences of expert instances
$j_1^T = (j_1, \ldots, j_T)$, where legality means that for all time instances $t$, the
considered expert $j_t$ is active (i.e., is in $E_t$). We call compound experts the elements of $\cL$.
Similarly, we denote by $\cC$ the set of all legal sequences of convex weight vectors
$\bq_1^T = (\bq_1, \ldots, \bq_T)$, where legality means that for all time instances $t$, the
considered convex weight vector $\bq_t$ puts positive masses only on elements in $E_t$.
We call compound convex weight vectors the elements of $\cC$.

For such compound experts $j_1^T$ or compound convex weight vectors $\bq_1^T$, we denote by
\[
\size \bigl( j_1^T \bigr) = \sum_{t=2}^T \ind_{ \{ j_{t-1} \ne j_t \} }
\qquad \mbox{and} \qquad
\size \bigl( \bq_1^T \bigr) = \sum_{t=2}^T \ind_{ \{ \bq_{t-1} \ne \bq_t \} }
\]
their numbers of switches (the number minus one of elements in the partition of
$\{ 1,\ldots,T \}$ into integer subintervals corresponding to the use
of the same expert or convex weight vector). For $0 \leq m \leq T-1$, we then respectively
define $\cL_m$ and $\cC_m$ as the subsets
of $\cL$ and of $\cC$ containing the compound experts and compound convex weight vectors with at most $m$ shifts.
When $m$ is too small, the subsets $\cL_m$ and $\cC_m$ might be empty.

The regrets of a rule $\cA$ with respect to $j_1^T \in \cL$ and $\bq_1^T \in \cC$
are respectively given by
\[
R_T\bigl( \cA,j_1^T \bigr) = \sum_{t=1}^T \Bigl( \ell_t(\bp_t) - \ell_t\bigl(\delta_{j_t}\bigr) \Bigr)
\qquad \mbox{and} \qquad
R_T\bigl( \cA, \bq_1^T \bigr) = \sum_{t=1}^T \bigl( \ell_t(\bp_t) - \ell_t(\bq_t) \bigr) \,.
\]
Since $\cL_m \subseteq \cC_m$ (up to the identification of expert indexes $j$ to convex weight vectors $\delta_j$),
it is more difficult to control the regret with respect to all elements of $\cC_m$
than the one with respect to simply $\cL_m$.

The aggregation rule presented in Figure~\ref{fig:FSrule} (when used directly on the losses $\ell_t$)
is actually nothing but an efficient computation
of the rule that would consider all compound experts and perform exponentially weighted averages
on them in the spirit of the rule $\cE_\eta$ but with a non-uniform
prior distribution. We will call it the fixed-share rule for specialized experts; we denote
it by $\cF_{\eta,\alpha}$ as it depends on two parameters, $\eta > 0$ and $0 \leq \alpha \leq 1$.
This rule is a straightforward adaptation to the setting of specialized experts
of the original fixed-share forecaster of~\cite{HeWa98}, see also \cite[Section~5.2]{CeLu06}.
\begin{figure}[t]
\emph{Parameters}: learning rate $\eta >0$ and mixing rate $0\leq \alpha\leq 1$
\smallskip\noindent

\emph{Initialization}: $(w_{1,0},\ldots,w_{N,0}) = \displaystyle{ \frac{1}{|E_1|} \bigl( \ind_{ \{ 1 \in E_1 \} }, \, \ldots,
\, \ind_{ \{ N \in E_1 \} } \bigr) }$
\smallskip\noindent

{\em For} each round $t = 1,2,\ldots,T$,
  \smallskip\noindent
  \begin{itemize}
  \item[(1)] predict $\displaystyle{\wh{y}_t = \frac{1}{\sum_{k=1}^N w_{k,t-1}} \sum_{j=1}^N w_{j,t-1} \, f_{j,t}}$\,;
  \item[(2)] [loss update] observe $y_t$ and define for each $i=1,\ldots,N$,
 \begin{numcases}{v_{i,t}= }
     \nonumber
 w_{i,t-1} \, e^{- \eta \ell_t(\delta_i)} & if $i \in E_t$, \\
     \nonumber
     \mbox{undefined} & if $i \notin E_t$;
  \end{numcases}
  \item[(3)] [share update] let $w_{j,t} = 0$ if $j \not\in E_{t+1}$ and
   \[
w_{j,t} =\frac{1}{\abs[E_{t+1}]}\sum_{i \in E_t\setminus E_{t+1}}\!\!\!v_{i,t}+ \frac{\alpha}{\abs[E_{t+1}]}\sum_{i \in E_t\cap E_{t+1}}\!\!\!v_{i,t}+(1-\alpha)\,\ind_{ \{ j\in E_t\cap E_{t+1} \} }\, v_{j,t}
   \]
   if $j \in E_{t+1}$, with the convention that an empty sum is null and denoting by $\abs[E_{t+1}]$ the cardinality of $E_{t+1}$.
  \end{itemize}
\rule{\linewidth}{.5pt}
\caption{\label{fig:FSrule} The fixed-share aggregation rule $\cF_{\eta,\alpha}$.}
\end{figure}

Its performance bound is stated below;
it follows from a straightforward but lengthy adaptation of the
techniques used in~\cite{HeWa98} and~\cite[Section~5.2]{CeLu06}.
We thus provide it in the appendix of this paper (Section~\ref{sec:app}), for the sake of completeness and to show how the share
update of Figure~\ref{fig:FSrule} was obtained.

\begin{theorem}
\label{prop:FSEWA}
We assume that the loss functions $\ell_t$ are convex and uniformly bounded;
we denote by $L$ a constant such that the quantities $\ell_t(\delta_i)$ all belong to $[0,L]$ when $i$ varies
in $E_t$ and $t$ varies from $1$ to $T$.
For all $m \in \{ 0,\ldots,T-1\}$, the regret of $\cF_{\eta,\alpha}$ is uniformly bounded over
all such sequences of expert forecasts and observations as
\begin{equation}
\label{eq:regretboundFSEWA}
\max_{j_1^T \in \cL_m} {R}_{T} \bigl( \cF_{\eta,\alpha}, j_1^T \bigr)
\leq \frac{m+1}{\eta} \ln N + \frac{1}{\eta}
\ln \frac{1}{\alpha^m \, (1-\alpha)^{T-m-1}} + \frac{\eta}{8} L^2 T\,.
\end{equation}
\end{theorem}

The (theoretically almost) optimal bound in the theorem above can be obtained by defining
the binary entropy $H$ as $H(x) = x \ln x + (1-x) \ln (1-x)$ for
$x \in [0,1]$, by fixing a value of $m$, and by carefully choosing parameters $\alpha^{\star}$
and $\eta^\star$ depending on $m$, $L$, and $T$:
\[
\max_{j_1^T \in \cL_m} {R}_{T} \bigl( \cF_{\eta^{\star},\alpha^{\star}}, j_1^T \bigr) \leq
L \sqrt{\frac{T}{2} \Bigl( (m+1) \ln N + (T-1) H \bigl( m/(T-1) \bigr) \Bigr)}\,,
\]
which is $o(T)$ as desired as soon as $m = o(T)$. Of course,
the theoretical optimal choices depend on $T$ and $m$, so that here also sequential adaptive choices
are necessary; see Section~\ref{sec:calibr} for a discussion. \medskip

By resorting to the gradient trick defined in Section~\ref{sec:subgrad}, i.e.,
by replacing the losses $\ell_t$ in the loss update of Figure~\ref{fig:FSrule} by the pseudo-losses $\tl_t$,
one obtains a variant of the previous forecaster, denoted by $\cF^{\grad}_{\eta,\alpha}$.
The following performance bound is a corollary of Theorem~\ref{prop:FSEWA};
a formal proof is provided in appendix (Section~\ref{sec:app3}).

\begin{corollary}
\label{cor:FSEG}
We assume that the loss functions $\ell_t$ are convex and (sub)differentiable over $\cX$,
with (sub)gradient functions uniformly bounded in the supremum norm as $t$ varies by $G$.
For all $m \in \{ 0,\ldots,T-1\}$, the regret of $\cF^{\grad}_{\eta,\alpha}$ is uniformly bounded over
all such sequences of observations and of expert forecasts as
\begin{equation}
\label{eq:regretboundFSEG}
\max_{\bq_1^T \in \cC_m} {R}_{T} \bigl( \cF^{\grad}_{\eta,\alpha}, \bq_1^T \bigr)
\leq \frac{m+1}{\eta} \ln N + \frac{1}{\eta}
\ln \frac{1}{\alpha^m \, (1-\alpha)^{T-m-1}} + \frac{\eta}{2} G^2 T\,.
\end{equation}
\end{corollary}

\subsection{Sequential automatic tuning of the parameters on data}
\label{sec:calibr}

The aggregation rules discussed above are only semi-automatic strategies,
as they rely on fixed-in-advance parameters $\eta$ (and possibly $\alpha$) that are not tuned on data.
Fully sequential aggregation rules need to set these parameters online.
Theoretically almost optimal ways of doing so exist; for instance,
\cite{AuCeGe02,CeMaSt07} indicate ways to online tune the learning rates $\eta$
for exponentially weighted average rules $\cE$ and $\cE^{\grad}$ so as to
achieve almost the same regret bounds as if the parameters $L$, $G$, and $T$
were known in advance. However, the learning rates thus obtained usually perform poorly in practice;
see~\cite{MaStMa09} for an illustration of this fact on different data sets.
The same is observed on our data sets (results not reported); this does not come as a surprise
as the theoretically optimal parameters $\eta^\star$ themselves perform poorly, see Remarks~\ref{rk:perftheo1}
and~\ref{rk:perftheo2} in the empirical studies. Therefore, in spite of the existence of theoretically satisfactory methods,
other ones need to be designed based on more empirical considerations.

We do so below but for the sake of completeness we discuss
first the symmetric case of the tuning of the parameter $\alpha$ of the fixed-share type rules.
These rules need actually to tune two parameters, $\eta$ and $\alpha$; the two tunings are equally important, as is illustrated by the
performance reported in Tables~\ref{tab:offlineslov2} and~\ref{tab:offlineFrench2}.
The tuning of $\eta$ could be done according to the same theoretical methods as mentioned above
(e.g., \citep{AuCeGe02,CeMaSt07}) but the same issues of practical performance arise. As for $\alpha$, it is possible in theory
not to tune it but to aggregate instances of the rule corresponding to different values of $\alpha$,
where these values lie in a thin enough grid; again, the rule performing this aggregation,
e.g., an exponentially weighted average rule, needs to be properly tuned as far as its learning rate
$\eta'$ is concerned. Such a double-layer aggregation was proposed by \cite{MoJa03}, see also~\cite{Dutch}.
We implemented it on our second data set and it turned out to have a performance similar
to the empirical method we detail now, as long as the learning rates $\eta$ and $\eta'$ were properly set both in the base
rules and in the second-layer aggregation, e.g., as follows.

\subsubsection*{An empirical online tuning of the parameters}

We describe the method in a general framework; it is
due to Vivien Mallet and was proposed in the technical report by~\cite{GeMaSt08} (but never published elsewhere
to the best of our knowledge).
Let $\cA_\lambda$ be a family of sequential aggregation rules relying each on some parameter
$\lambda$ (possibly vector-valued) taking its values in some set $\Lambda$.
Given the past observations and the past and present
forecasts of the experts, the rule index by $\lambda$ prescribes at time instance $t$ a convex weight vector
which we denote by $\bp_t \bigl( \cA_\lambda \bigr)$.

The weights used by the fully sequential aggregation
rule based on the family of rules $\cA_\lambda$, where $\lambda \in \Lambda$,
will be denoted by $\wh{\bp}_t$.
We assume that the considered family is such that
$\bp_1 \bigl( \cA_\lambda \bigr)$ is independent of $\lambda$, so that
$\wh{\bp}_1$ equals this common value. Then, at time instances $t \geq 2$,
\begin{equation}
\label{eq:calibr}
\wh{\bp}_t = \bp_t \Bigl( \cA_{ \wh{\lambda}_{t-1} } \Bigr)
\qquad \mbox{where} \qquad
\wh{\lambda}_{t-1} \in \argmin_{\lambda \in \Lambda} \,\, \sum_{s=1}^{t-1} \ell_s \Bigl( \bp_s \bigl(
\cA_\lambda \bigr) \Bigr)\,;
\end{equation}
that is, we consider, for the prediction of the next time instance, the aggregated forecast
proposed by the best so far member of the family of aggregation rules.
Because of this formulation, we will speak of a meta-rule in the sequel.
We can however offer no theoretical guarantee for the performance
of the meta-rule in terms of the performance of the underlying family. \medskip

Computationally speaking, we need to run in parallel all the instances of $\cA_\lambda$,
together with the meta-rule. This of course is impossible as soon as $\Lambda$ is not finite;
for the families considered above we had $\Lambda = (0,+\infty)$ and $\Lambda = (0,+\infty) \times [0,1]$.
This is why, in practice, we only consider a finite grid $\wt{\Lambda}$
over $\Lambda$ and perform the minimization of (\ref{eq:calibr}) only on
the elements of $\wt{\Lambda}$ instead of performing it on the whole set $\Lambda$.
A final choice still seems to be left to the user, namely, how to design this finite grid $\wt{\Lambda}$.
For the first data set (in Section~\ref{sec:onlc-Slov}) we fix it somewhat arbitrarily. Based on the
observed behaviors, we then propose for the second data set (in Section~\ref{sec:fulladapt})
a way to construct online the grid $\wt{\Lambda}$, finally leading to a fully sequential meta-rule.

\subsubsection*{Literature review of empirical studies in our framework}

Several articles report applications of prediction based on expert
advice to real data. They do not investigate the online tuning issues discussed above
and can be clustered into three categories as far as the tuning of the parameters is concerned
(there is often only a learning rate $\eta$ to be set).

The first group chooses in the experiments the theoretically optimal parameters (sometimes,
for instance, in the case of square losses, these are
given by the rates $\eta$ such that a property of exp-concavity holds). This would be possible
as well in our context with improved regret bounds but only for the basic versions of our forecasters,
not for their gradient versions (which will be seen to obtain a much improved performance in practice). Furthermore,
even such choices of $\eta$ are slightly suboptimal on our data sets with respect to the fully sequential
tuning described above. Actually, tuning $\eta$ in such a way,
one only targets the performance of the best expert, not the one of the best
convex combination of the experts (which is significantly better). Examples
of such articles and fields of application include the management of the tradeoff between energy consumption and performance in wireless networks
(\citep{MoJa03}), the tracking of climate models (\citep{MoCIDU,Jac11}), the network traffic demand (\citep{Network}),
the prediction of GDP data (\citep{Jac11}), and also the online aggregation of portfolios (e.g., \citep{Cov91,StLu05},
but the literature is vast).
In particular, as far as the latter application is concerned, we note that \cite{Gog00}
indicates that the studied forecasters do not differ significantly from uniform averages of the experts;
this is because the parameter $\eta$ is not set large enough. This is why we designed a method to tune
it automatically based on past data to get the right scale of the problem.

The second group of articles only reports results of optimal-in-hindsight parameters (and sometimes argues that
the performance is not very sensitive to the tuning, a fact that we do not observe on the data sets studied
in this paper). The studied topics are, for instance, the forecasting of air quality (\citep{MaStMa09,Mal10})
and the prediction of outcomes of sports games (\citep{DaMaPeSaGa06}).

The third group reports the performance of various values of the parameters
without choosing between them in advance, for instance, \cite{VoZh08}
for the latter application or \cite{StLu05} already mentioned above.

\newpage
\section{Methodology followed in the empirical studies}
\label{sec:methodo}

We provide a standardized outline of the treatment of the two data sets discussed in the next sections.

\bookbox{
\emph{Outline of the empirical studies of performance of the sequential aggregation rules}
\begin{enumerate}
\item Design experts (based on some historical data). \vspace{.1cm}
\item Choose a loss function and evaluate the performance of the experts (on new data). \vspace{.1cm}
\item For each family of strategies compute the performance corresponding to the best constant choices of the parameters in hindsight. \vspace{.1cm}
\item Assess the quality of the operational performance, i.e., the performance obtained after some automatic and sequential tuning (see Section~\ref{sec:calibr}). \vspace{.1cm}
\item Provide additional results and comments (e.g., a robustness study).
\end{enumerate}
}

By evaluation of the performance of the experts we mean the assessment of the accuracy
obtained by some simple strategies like the uniform average of the forecasts of the active experts (a
strategy easily implementable online) or by some oracles, like the best single expert or the best
constant convex combination of the experts. Finally, the so-called prescient strategy is the strategy that picks at each time instance
the best forecast output by the set of experts; it indicates a bound on the performance
that no aggregation strategy can improve on given the data set (given the expert forecasts and the observations).
It corresponds to the best element in $\cL_{T-1}$.

\section{A first data set: Slovakian consumption data}
\label{sec:slov}

The data was provided by the Slovakian
subbranch of the French electricity provider EDF.
It is formed by the hourly predictions of 35 experts and the corresponding
observations (formed by hourly mean consumptions) on the period from January~1, 2005 to December~31, 2007.
In this part and unlike for the French data set of the next part, we
have absolutely no information on how the experts were built and we merely consider them as
black boxes.

\begin{figure}[p]
\begin{center}
\includegraphics[width=7.5cm]{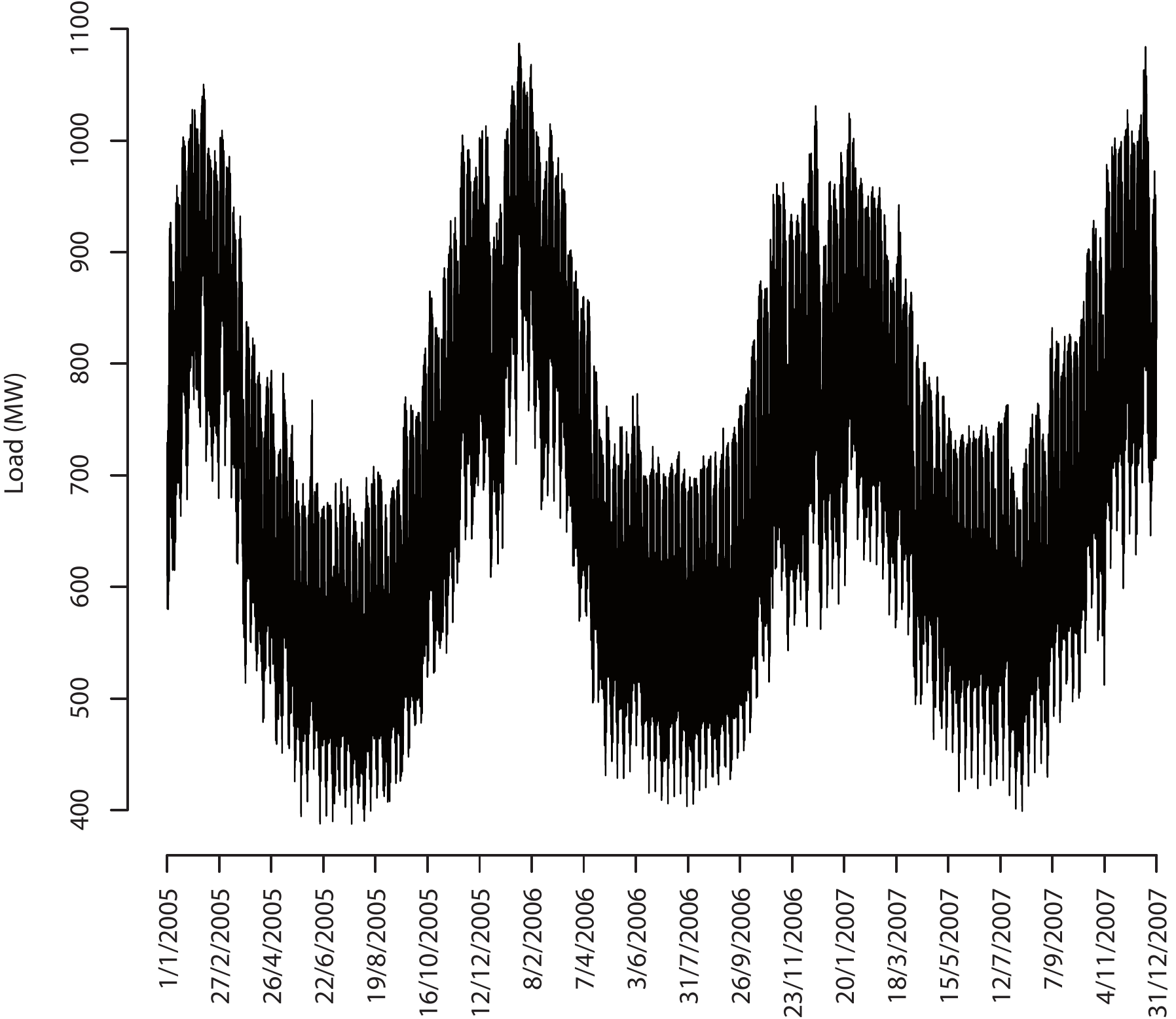}
\end{center}
\caption{\label{fig:plotload}
The observed hourly electricity consumptions encountered by the Slovakian
subbranch between January 1, 2005 and December 31, 2007.}
\end{figure}

As the behavior of electricity consumption depends heavily on the hour of the day and the data set is large enough, we
parsed it set into 24 subsets (one per hour interval of the day) and only report
the results obtained for one-day-ahead prediction
on a given (somewhat arbitrarily chosen) hour interval: the interval 11:00--12:00.
The characteristics of the observations $y_t$ of this hour frame are described in Table~\ref{tab:charslov}
while all observations (for all hour frames) are plotted in
Figure~\ref{fig:plotload}.

The considered loss function is the square loss and we will not report cumulative losses
but root mean square errors ({\rmse}), i.e., roots of the per-round cumulative losses. For instance,
for a given convex combination $\bq \in \cX$,
\[
\rmse(\bq) = \sqrt{\frac{1}{{\sum_{t=1}^T \bq(E_t) }}
\sum_{t=1}^T \left( \sum_{j \in E_t} q_j^{E_t} f_{j,t} - y_t \right)^{\! 2} \bq(E_t) }\,
\]
while for an aggregation rule $\cA$,
\[
\rmse(\cA) = \sqrt{\frac{1}{T} \sum_{t=1}^T \bigl( \wh{y}_t - y_t \bigr)^2}\,.
\]

In this section, we will omit the unit MW (megawatt) of the observations and predictions of the electricity
consumption, as well as the one of their corresponding $\rmse$.

\begin{table}[p]
\caption{\label{tab:charslov} Some characteristics of the observations $y_t$ (hourly mean consumptions)
of the Slovakian data set for the time intervals 11:00--12:00.}
{\small
\begin{center}
\begin{tabular}{lc}
\hline
Number of days $D$ & 1\,095 \\
Time intervals & Only 11:00--12:00 \\
\hline
Number of instances $T$ & 1\,095 ($= 1\,095 \times 1$) \\
Number of experts $N$ & 35 \\
\hline
Unit & MW \\
Median of the $y_t$ & 702.6 \\
Bound $B$ on the $y_t$ & 1020.0 \\
\hline
\end{tabular}
\end{center}
}
\end{table}

\begin{figure}[p]
\begin{tabular}{cc}
\includegraphics[width=5.5cm]{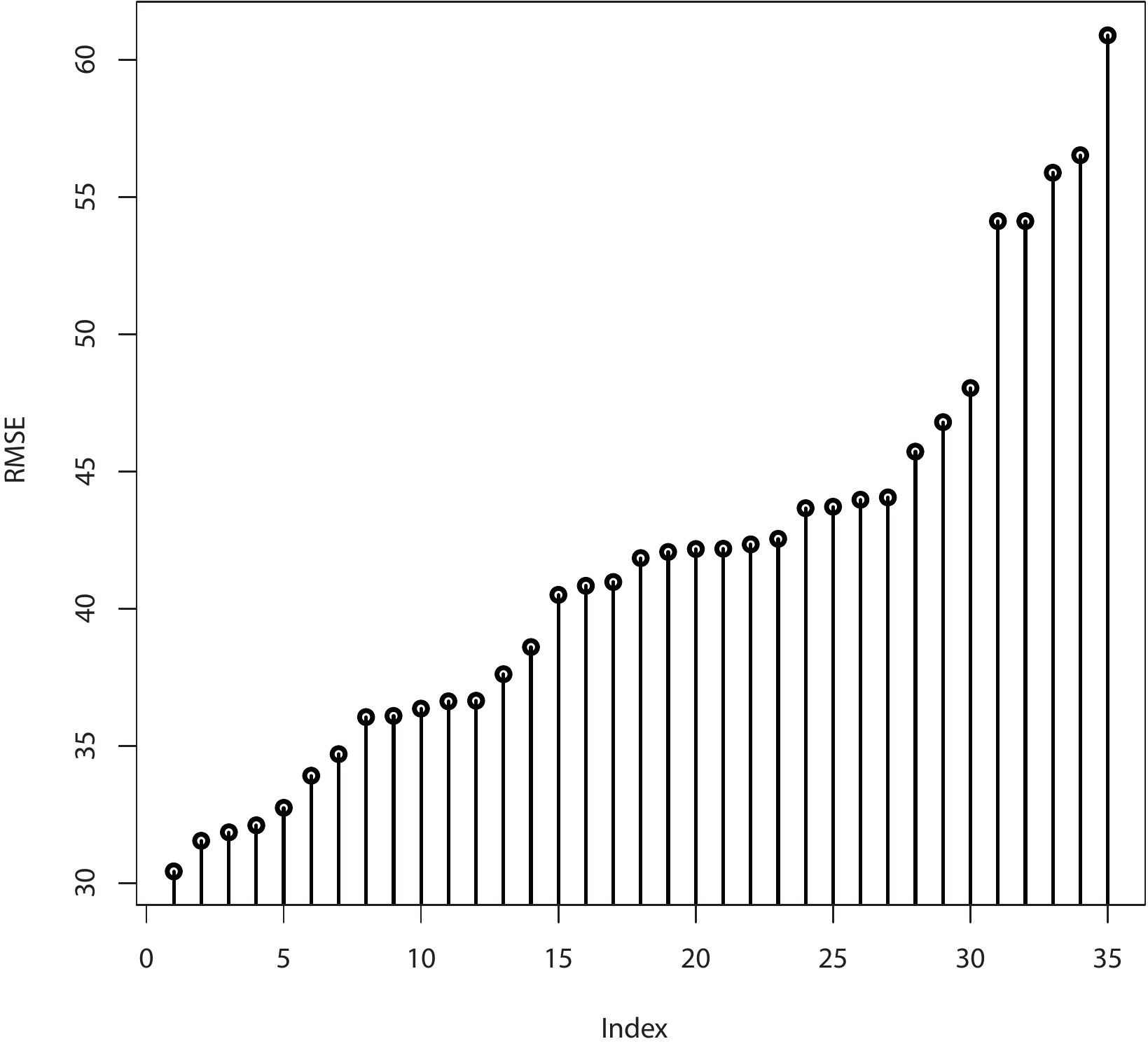} &
\includegraphics[width=5.5cm]{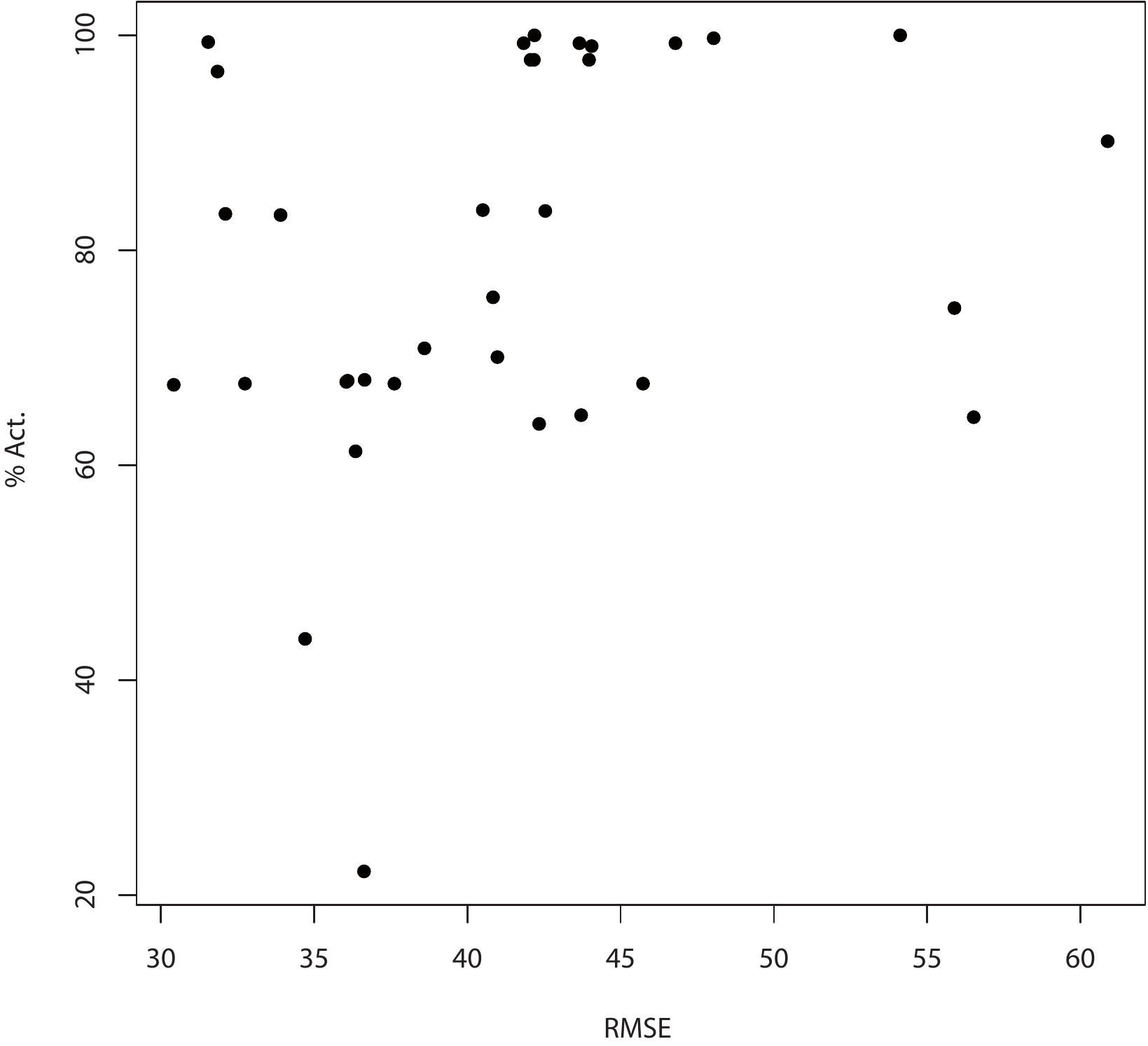}
\end{tabular}
\caption{\label{fig:charslov} Graphical representations of the performance of the experts
of the Slovakian data set: sorted $\rmse$ (left) and
$\rmse$--frequency of activity pairs (right).}
\end{figure}

\subsection{Benchmark values: performance of the experts and of some oracles}
\label{sec:benchm-Slov}

The characteristics of the experts are depicted in Figure~\ref{fig:charslov}.
The bar plot represents the values of the $\rmse$ of the 35 available experts.
The scatter plot relates the $\rmse$ of each of the expert to its frequency of activity,
that is, it plots the pairs, for all experts $j$,
\begin{equation}
\label{eq:rmseact}
\left( \rmse(j), \ \frac{\sum_{t=1}^T \ind_{ \{ j \in E_t \} }}{T} \right)\,.
\end{equation}

We present in Table~\ref{tab:oracles} the values\footnote{All of them
have been computed exactly, except the ones that involve
minimizations over simplexes of convex weights, for which
a Monte-Carlo stochastic approximation method was used.
}
of the {\rmse} of several procedures, all of them but the first two being oracles.
The procedure $\cU$ is an aggregation rule that simply chooses at each time instance $t$ the uniform convex
weight vector on $E_t$. Its {\rmse} differs from the one of the uniform
convex weight vector $(1/35,\ldots,1/35)$ as the {\rmse} of the latter gives a weight
to each instance $t$ that depends on the cardinality of $E_t$.

The fact that the {\rmse} of the best compound expert with size at most 10
is larger than the {\rmse} of the best single expert is explained by the fact
that some overall good experts refrain from predicting at some time instances when
all active experts perform poorly, while compound experts are required to output
a prediction at each time instance. The fact that such good experts tend not to
form predictions at instances that are more difficult to cope with can also be
seen from the fact that $\rmse(\cU)$ is larger than $\rmse \bigl( (1/35,\ldots,1/35) \bigr)$,
since the second uniform average rule is evaluated with unequal weights put on
the different time instances (more weight put on instances when more experts are active).

A final series of oracles is given by
partitioning time into subsets of instances with constant sets of active experts;
that is, by defining
\[
\bigl\{ E^{(1)}, \ldots, E^{(K)} \bigr\} = \bigl\{ E_t, \,\, t \in \{ 1,\ldots,T \} \bigr\}
\]
and by partitioning time according to the values $E^{(k)}$ taken by the sets of active experts $E_t$.
The corresponding natural oracles are
\begin{equation}
\label{eq:Bpart}
\min \ \left\{ \,\, \sqrt{\frac{1}{T} \sum_{k = 1}^K \ \sum_{t : E_t = E^{(k)}} \bigl( f_{j^k,t} - y_t \bigr)^2} \, ,
\ \ \mbox{with} \ j^k \in E^{(k)} \ \ \mbox{for all $k = 1,\ldots,K$} \right\}\,,
\end{equation}
which corresponds to the choice of the best expert on each element of the partition,
and
\begin{multline}
\label{eq:BpartX}
\min \left\{ \sqrt{\frac{1}{T} \sum_{k = 1}^K \ \sum_{t : E_t = E^{(k)}} \left( \sum_{j \in E^{(k)}}
q^{(k)}_j f_{j,t} - y_t \right)^2} \, , \right. \\
\mbox{with} \ \bq^{(k)} \ \mbox{a convex weight vector on $ E^{(k)}$} \ \ \mbox{for all $k = 1,\ldots,K$} \biggr\}\,,
\end{multline}
which corresponds to the choice of the best convex weight vector on each element of the partition.
Even if there are relatively many elements in this partition, namely, $K = 74$,
the gain with respect to constant choices throughout time exists ({\rmse} of
29.1 versus 30.4 and 24.5 versus 29.2) but is less significant than the one achieved with compound
experts (which achieve a smaller {\rmse} of 23.1 already with a size $m = 50$).

\begin{table}[t]
\caption{\label{tab:oracles}
Definition and performance of several (possibly off-line) benchmark
procedures on the Slovakian data set; they serve as comparison points for on-line procedures.}
{\small
\begin{center}
\begin{tabular}{lcr}
\hline
Name of the benchmark procedure & Formula & Value \\
% \hline
% & & \\
\hline
Uniform sequential aggregation rule & $\rmse(\cU)$ \vspace{.1cm} & $= 31.1$ \\
Uniform convex weight vector & $\displaystyle{\rmse \bigl( (1/35,\ldots,1/35) \bigr)}$ \vspace{.1cm} & $= 30.7$ \\
Best single expert & $\displaystyle{\min_{j = 1,\ldots,35} \ \rmse(j)}$ & $= 30.4$ \\
Best convex weight vector & $\displaystyle{\min_{\bq \in \cX} \ \rmse(\bq)}$ & $= 29.2$ \\
\hline
% & & \\
% \hline
Best compound expert & & \vspace{.15cm}\\
Size at most $m = 10$ & $\displaystyle{\min_{j_1^T \in \cL_{10}} \ \rmse \bigl( j_1^T \bigr)}$ & $= 32.1$ \\
Size at most $m = 50$ & $\displaystyle{\min_{j_1^T \in \cL_{50}} \ \rmse \bigl( j_1^T \bigr)}$ & $= 23.1$ \\
Size at most $m = 200$ & $\displaystyle{\min_{j_1^T \in \cL_{200}} \ \rmse \bigl( j_1^T \bigr)}$ & $= 15.2$ \\
Prescient strategy (size at most $m = T - 1 = 1\,094$) & $\displaystyle{\min_{j_1^T \in E_1 \times E_2 \times \ldots \times E_T}
\ \rmse \bigl( j_1^T \bigr)}$ & $= \ \, 9.4$ \\
\hline
% & & \\
% \hline
On the $K = 74$ elements of a partition of time & \vspace{-.1cm} \\
according to the values of the active sets $E_t$ & \vspace{.1cm} \\
Best expert on each element & See (\ref{eq:Bpart}) & $= 29.1$ \\
Best convex weight vector on each element & See (\ref{eq:BpartX}) & $= 24.5$ \\
\hline
\end{tabular}
\end{center}
}
\end{table}

\subsection{Results obtained with constant values of the parameters}
\label{sec:perfSlovCst}

We now detail the practical performance of the sequential aggregation rules introduced in Section~\ref{sec:theory},
for fixed values of the parameters $\eta$ and $\alpha$ of the rules.
We report for each rule the best performance obtained; the corresponding parameters
are said the best constant choices in hindsight.
The performance of the families $\cE_\eta$, $\cE_\eta^{\grad}$, and $\cS_\eta^{\grad}$ is summarized in Table~\ref{tab:offlineslov1}.
We note that $\cE_\eta^{\grad}$ and $\cS_\eta^{\grad}$, when tuned with the best parameter $\eta$ in hindsight,
outperform their comparison oracle, the best convex weight vector (with a relative improvement of $3\,\%$ in terms of the {\rmse}),
while the performance of the best $\cE_\eta$ comes very close to the one of its respective comparison oracle, the best single expert ({\rmse}
of 30.4 versus 30.5). The performance of the fixed-share type rules $\cF_{\eta,\alpha}$ and $\cF_{\eta,\alpha}^{\grad}$
is reported in Table~\ref{tab:offlineslov2}.

\begin{remark}
\label{rk:perftheo1}
As in \cite{MaStMa09}, the best constant choices in hindsight are far away from the theoretically
optimal ones, given by $\eta^\star \approx 8 \times 10^{-8}$ for $\cE_\eta$,
$\eta^\star \approx 4 \times 10^{-8}$ for $\cS^{\grad}_\eta$, and $\eta^\star \approx 2 \times 10^{-8}$ for $\cE_\eta^{\grad}$.
\end{remark}

\begin{figure}[p]
\begin{tabular}{cc}
\includegraphics[height=4.8cm]{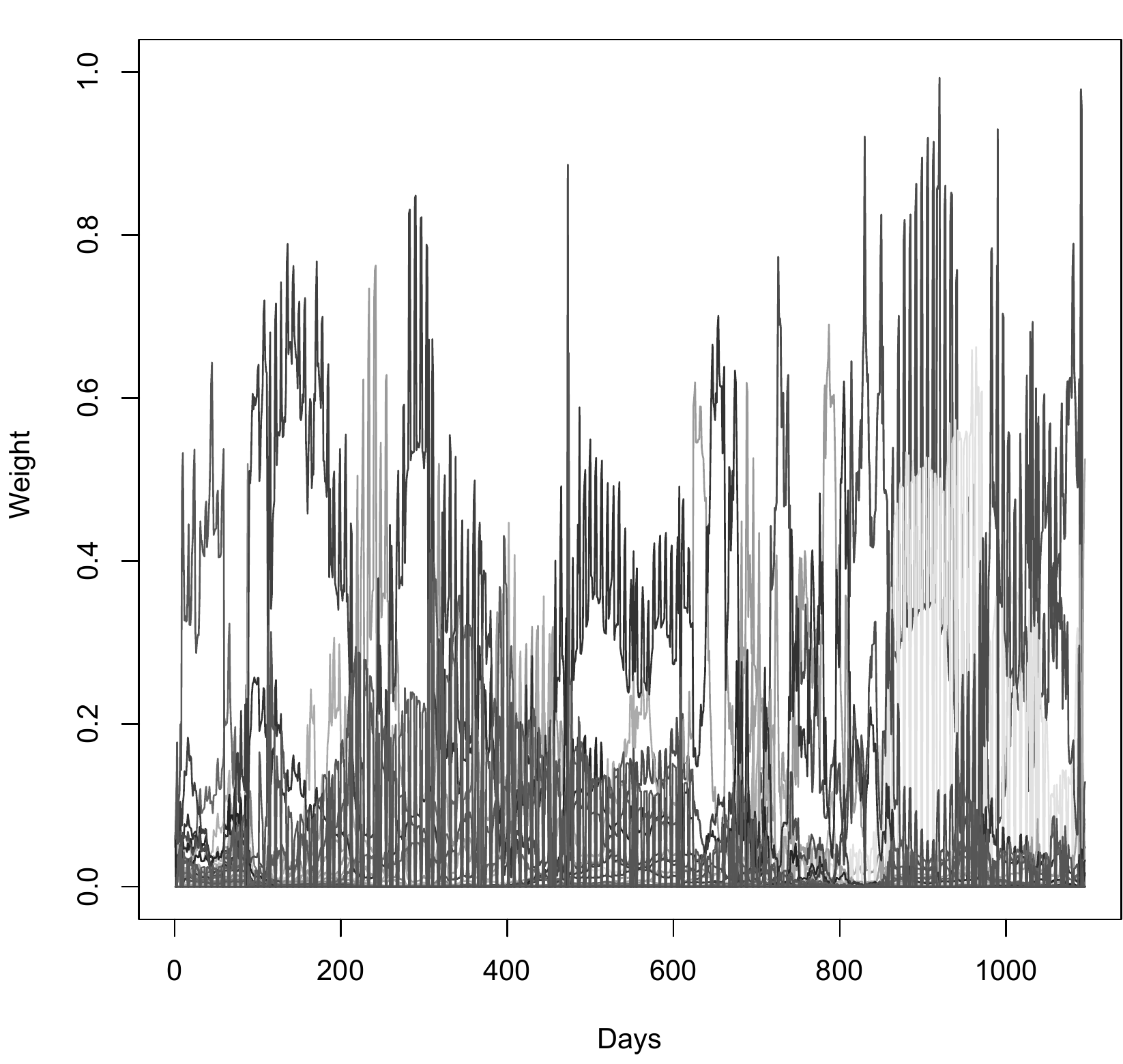}
&
\includegraphics[height=4.8cm]{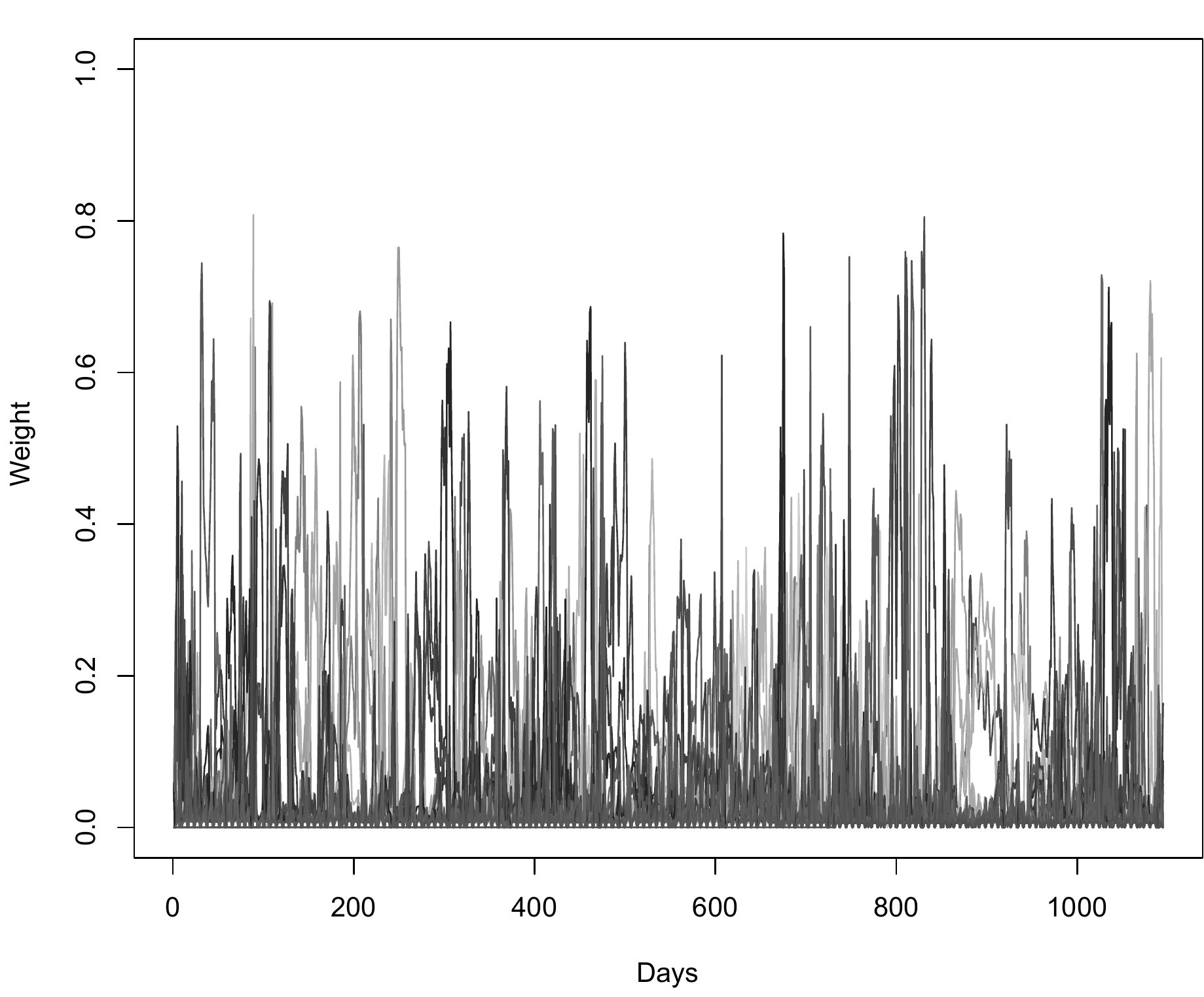}
\end{tabular}
\caption{\label{fig:weightsslov}
Graphical representations of the convex weights
associated at each time instance with the 35 experts by the rules $\cE_{10^{-4}}^{\grad}$ (left) and
$\cF_{2 \times 10^{-3},0.2}$ (right).}
\end{figure}

We close this preliminary review of performance by showing in Figure~\ref{fig:weightsslov}
that the considered rules fully exploit
the whole set of experts and do not concentrate on a limited subset of the experts. They carefully
adapt their convex weights as time evolves and remain reactive to changes of performance; in particular,
the sequences of weights do not converge to a limit vector.

\subsection{Results obtained with an online tuning of the parameters}
\label{sec:onlc-Slov}

\begin{figure}[p]
\begin{tabular}{cc}
\includegraphics[height=4.8cm]{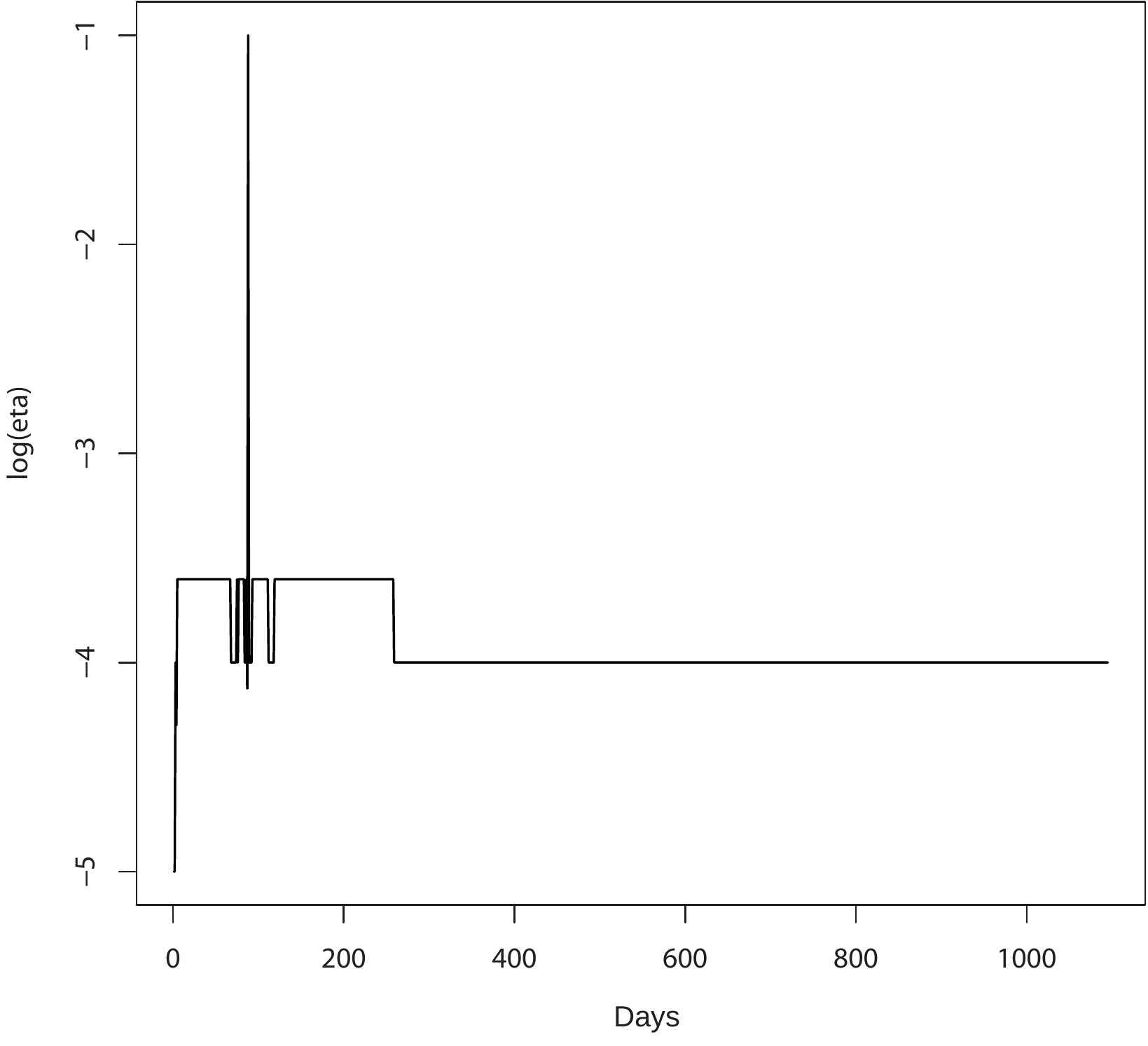}
&
\includegraphics[height=4.8cm]{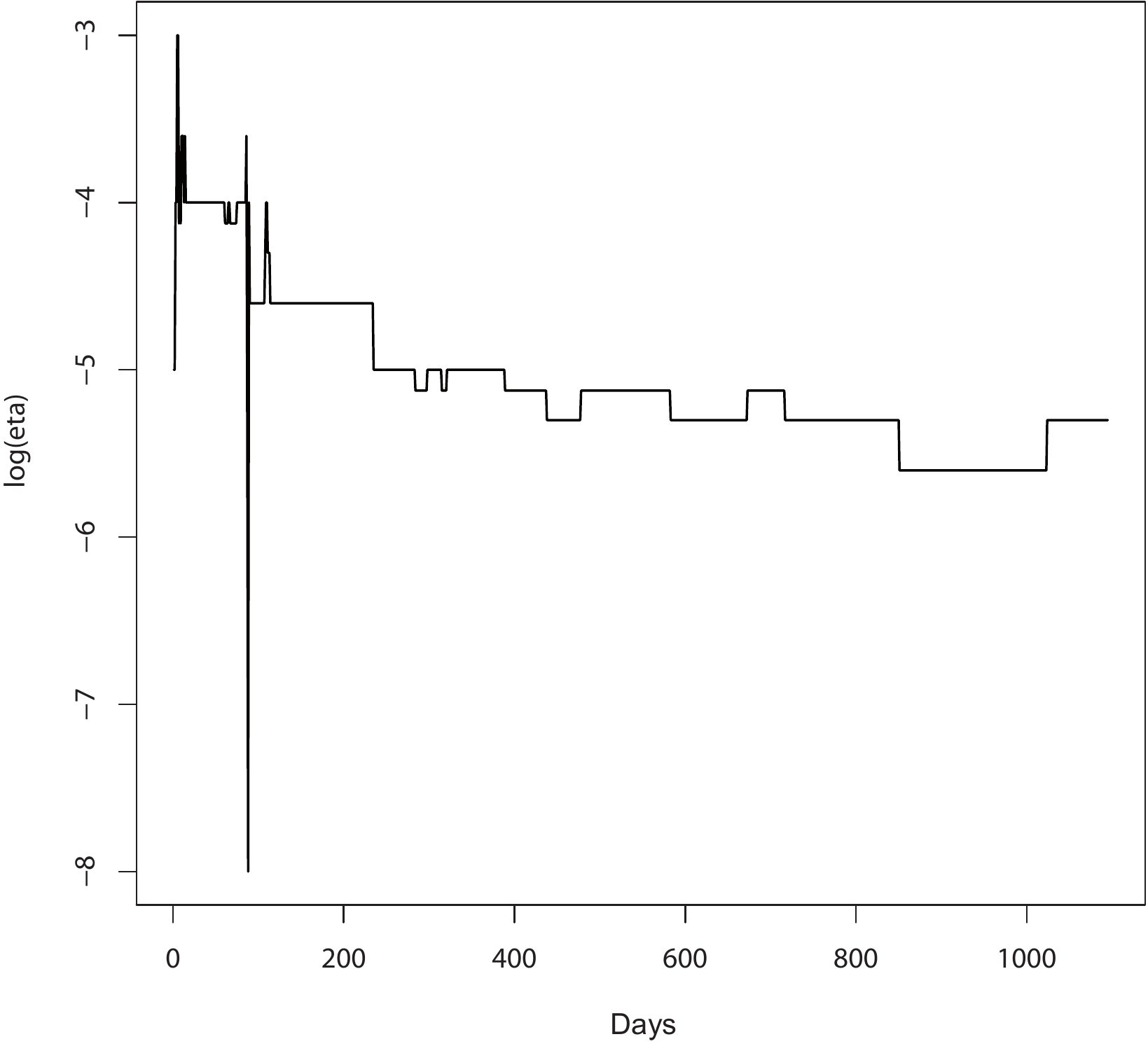}
\end{tabular}
\caption{\label{fig:etaonline}
Graphical representations of the sequences of
tuning parameters $\eta$ chosen by the meta-rule
selecting sequentially the values on the grid $\wt{\Lambda}_{\ell}$;
the base rules are $\cE_\eta^{\grad}$ (left) and $\cE_\eta$ (right).}
\end{figure}

We show in this section how the meta-rules constructed in Section~\ref{sec:calibr} can
get performance close to the one of the rules based on the best constant parameters in hindsight;
we do so, for this data set only, by fixing somewhat arbitrarily the used grids. Based on the observed
behaviors we then indicate for the second data set (in Section~\ref{sec:fulladapt}) how these grids
can be constructed online. For the exponentially weighted average rules $\cE_\eta$ and
$\cE_\eta^{\grad}$, the order of magnitude of the optimal values
$\eta^{\star}$ being around $10^{-8}$, we considered two finite grids for the tuning of $\eta$,
both with endpoints $10^{-8}$ and $1$: a smaller grid, with 9 logarithmically evenly
spaced points,
\[
\wt{\Lambda}_s = \bigl\{ 10^{-k}, \ \mbox{for} \ k \in \{ 0,1,\ldots,8 \} \bigr\}\,,
\]
and
a larger grid, with 25 logarithmically evenly spaced points,
\[
\wt{\Lambda}_{\ell} = \bigl\{ m \times 10^{-k}, \ \mbox{for} \ k \in \{ 1,\ldots,8 \} \ \mbox{and} \
m \in \{ 1, \, 2.5, \, 5 \} \bigr\} \cup \{ 1 \} \,.
\]
The performance on these grids with respect to the best constant choice
of $\eta$ in hindsight is summarized in Table~\ref{tab:onlineslov1}. We note that the good performance
obtained for the best choices of the parameters in hindsight is preserved
by the adaptive meta-rules resorting to the grids. The sequences
of choices of $\eta$ on the largest grid $\wt{\Lambda}_{\ell}$
are depicted in Figure~\ref{fig:etaonline}.

\begin{figure}[p]
\begin{tabular}{cc}
\includegraphics[height=4.8cm]{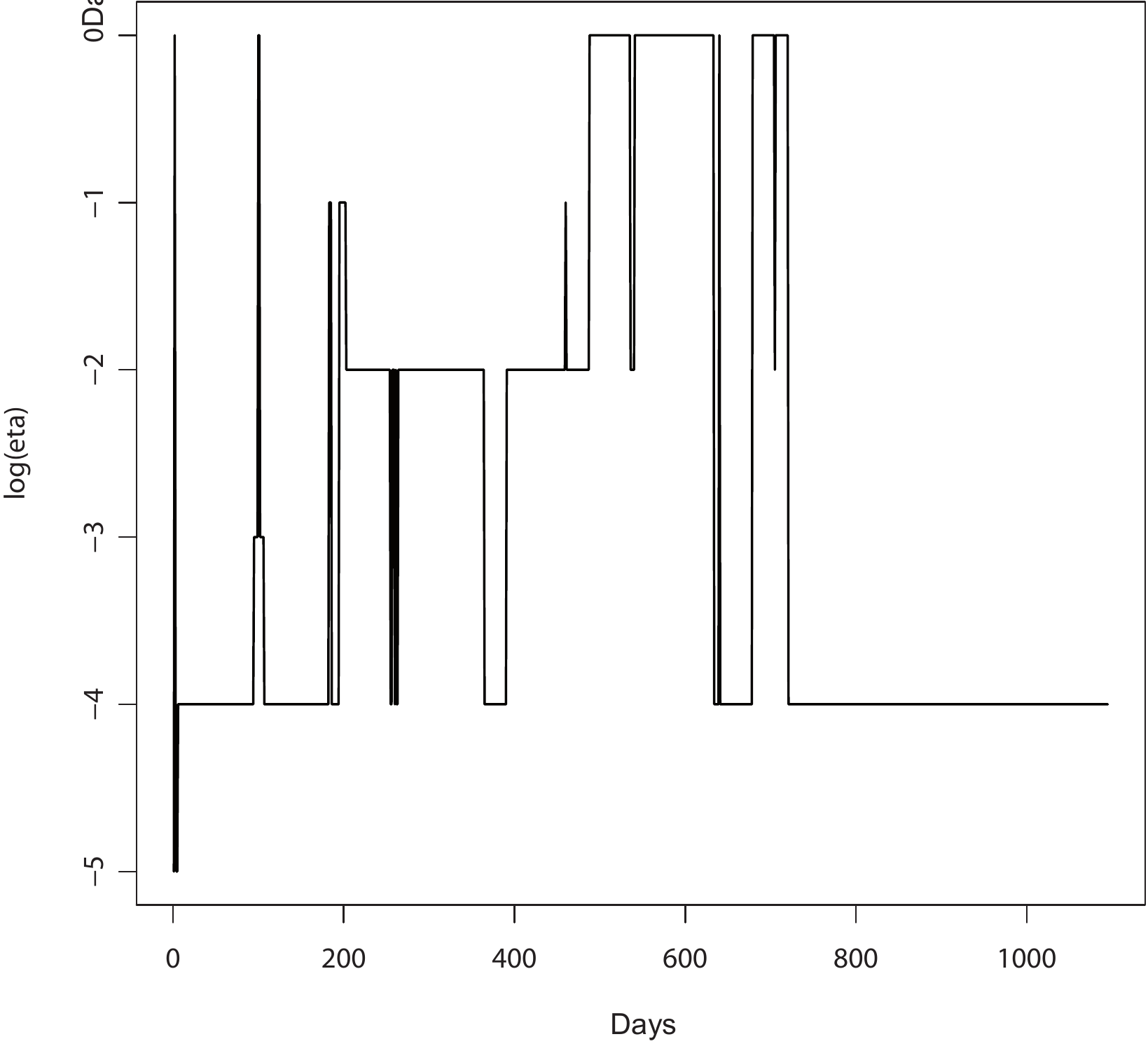}
&
\includegraphics[height=4.8cm]{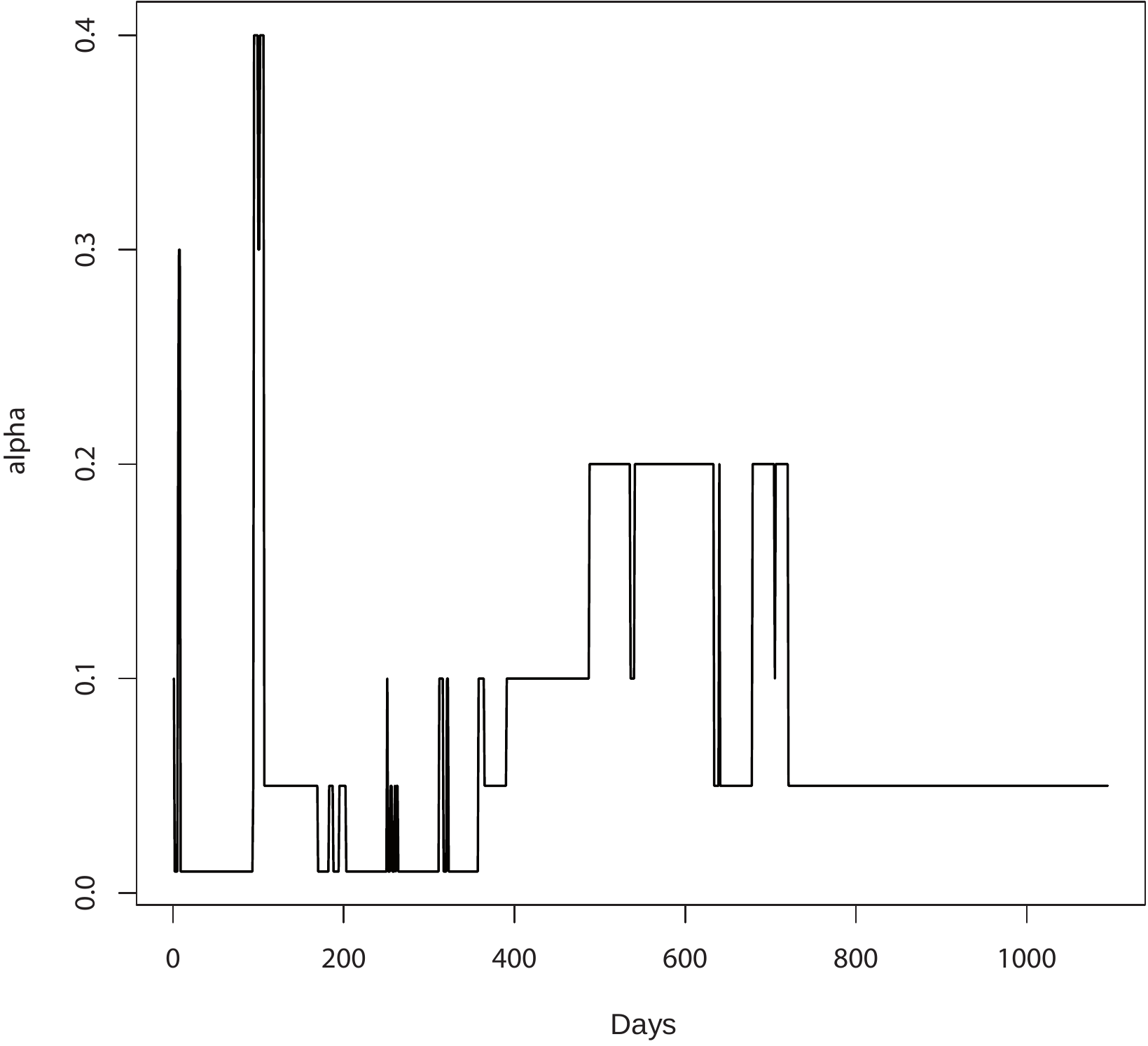}
\end{tabular}
\caption{\label{fig:etaalphaonline}
Graphical representations of the sequences of
tuning parameters $\eta$ (left) and $\alpha$ (right) chosen by the meta-rule
selecting sequentially the values on the grid $\wt{\Lambda}_{\mbox{\tiny FS}}$;
the base rule is $\cF_{\eta,\alpha}^{\grad}$.}
\end{figure}

For the fixed-share type rules $\cF_{\eta,\alpha}$ and
$\cF_{\eta,\alpha}^{\grad}$, two parameters have to be tuned: we need to take a finite grid in $\Lambda = (0,+\infty) \times [0,1]$,
e.g., similarly to above,
\[
\wt{\Lambda}_{\mbox{\tiny FS}} = \bigl\{
(10^{-k}, \, \alpha), \ \mbox{for} \ k \in \{ 0, 1, \ldots, 8 \} \ \mbox{and}
\ \alpha \in \{ 0.01, \, 0.05, \, 0.1, \, 0.2, \, 0.3, \, 0.4 \} \bigr\}\,.
\]
The performance on this grid is summarized in Table~\ref{tab:onlineslov2}
while the sequences
of choices of $\eta$ and $\alpha$ on the grid $\wt{\Lambda}_{\mbox{\tiny FS}}$
are depicted in Figure~\ref{fig:etaalphaonline}.
The same comments as above on the preservation of the good performance apply.

\begin{table}[p]
\caption{\label{tab:offlineslov1}
Performance obtained by the sequential aggregation rules
$\cE_\eta$, $\cE_\eta^{\grad}$, and $\cS^{\grad}_\eta$ for various choices of
$\eta$; the smallest {\rmse} obtained for each rule is underlined.}
{\small
\begin{center}
\begin{tabular}{ll*{8}{c}}
\hline
Value of & $\eta$ \ \ & $10^{-8}$ & $10^{-7}$ & $10^{-6}$ & $4 \times 10^{-6}$ & $10^{-5}$ & $10^{-4}$ & $10^{-3}$ \\
\hline
{\rmse} of & $\cE_\eta$ & 31.3 & 31.2 & 30.8 & \underline{30.5} & 30.9 & 32.7 & \\
           & $\cE_\eta^{\grad}$ & & 31.3 & 30.9 & & 29.8 & \underline{28.2} & 33.5 \\
           & $\cS^{\grad}_\eta$ & & 31.3 & 30.9 & & 29.8 & \underline{28.2} & 34.7 \\
\hline
\end{tabular}
\end{center}
}
\end{table}

\begin{table}[p]
\caption{\label{tab:offlineslov2}
Performance obtained by the sequential aggregation rules
$\cF_{\eta,\alpha}$ and $\cF_{\eta,\alpha}^{\grad}$
for various choices of $\eta$ and $\alpha$;
the smallest {\rmse} obtained for each rule is underlined.}
{\small
\begin{center}
\begin{tabular}{ll*{9}{c}}
\hline
Value \hspace{-.2cm} & $\eta$ & $10^{-4}$ & $10^{-4}$ & $10^{-3}$ & $10^{-3}$ & $10^{-2}$ & $10^{-2}$ & \ & $2 \times 10^{-4}$ & $2 \times 10^{-3}$ \\
of    \hspace{-.2cm} & $\alpha$ & 0.05 & 0.2 & 0.1 & 0.2 & 0.05 & 0.2 & & 0.07 & 0.2 \\
\hline
{\rmse} \hspace{-.2cm} & $\cF_{\eta,\alpha}$ & 29.3 & 29.5 & 27.5 & 27.2 & 28.0 & 27.8 & & & \underline{27.0} \\
of      \hspace{-.2cm} & $\cF_{\eta,\alpha}^{\grad}$ & 28.0 & 28.9 & 29.3 & 29.2 & 28.7 & 28.5 & & \underline{27.2} & \\
\hline
\end{tabular}
\end{center}
}
\end{table}

\begin{table}[p]
\caption{\label{tab:onlineslov1}
Performance obtained by
the rules $\cE_\eta$ and $\cE_\eta^{\grad}$
for the best constant choice of $\eta$ in hindsight (left)
and when used as keystones of a meta-rule selecting
sequentially the values of $\eta$ on the chosen grids
(middle and right).}
{\small
\begin{center}
\begin{tabular}{ll*{3}{c}}
\hline
& & Best constant $\eta$ & Grid $\wt{\Lambda}_s$ & Grid $\wt{\Lambda}_{\ell}$ \\
\hline
{\rmse} of & $\cE_\eta$ & 30.5 & 31.1 & 30.7 \\
           & $\cE_\eta^{\grad}$ & 28.2 & 28.2 & 28.4 \\
\hline
\end{tabular}
\end{center}
}
\end{table}

\begin{table}[p]
\caption{\label{tab:onlineslov2}
Performance obtained by
the rules $\cF_{\eta,\alpha}$ and $\cF_{\eta,\alpha}^{\grad}$
for the best constant choices of $\eta$ and $\alpha$ in hindsight (left)
and when used as keystones of a meta-rule selecting
sequentially the values of $\eta$ and $\alpha$ on the grid
$\wt{\Lambda}_{\mbox{\tiny FS}}$ (right).}
{\small
\begin{center}
\begin{tabular}{llcc}
\hline
& & Best constant pair $(\eta,\alpha)$ & Grid $\wt{\Lambda}_{\mbox{\tiny FS}}$ \\
\hline
{\rmse} of & $\cF_{\eta,\alpha}$ & 27.0 & 27.8 \\
           & $\cF_{\eta,\alpha}^{\grad}$ & 27.2 & 28.5 \\
\hline
\end{tabular}
\end{center}
}
\end{table}

\clearpage

\clearpage

\section{A second data set: Operational forecasting on French data}
\label{sec:French}

The data set used in this part is a standard data set used by EDF R{\&}D department.
It contains the observed electricity consumptions as well as some side information, which consists of all the features that were shown to have a strong effect on electricity load; see,
e.g., \cite{Bu85}. Among others, one can cite
seasonal effects (most importantly, the seasonal variations of day lengths),
calendar events like vacation periods or public holidays, weather conditions (temperature, cloud cover, wind),
and weekly patterns of days.
We summarize below some of its characteristics and we refer the interested reader to~\cite{Dor08}
for a more detailed description.

It is divided into two sets. The first set ranges from September 1, 2002 to August 31, 2007.
We call it the estimation set and use it to design
the experts, which then provide forecasts throughout the period corresponding to the
second set. This second set covers the period from September 1, 2007 to August 31, 2008. We call it the validation set
and use it to evaluate the performance of the considered aggregation rules.
Actually, we exclude some special days from the validation set. Out of the 366 days between September 1, 2007 and August 31, 2008,
we keep 320 days. The excluded days correspond to public holidays (the day itself, as well as the days before and after it),
daylight saving days and winter holidays (that is, the period between December 21, 2007 and January 4, 2008);
however, we include the summer break (August 2008) in our analysis as we have access to
experts that are able to produce forecasts for this period.
The characteristics of the observations $y_t$ of the validation set (formed by half-hourly mean consumptions)
are described in Table~\ref{tab:charFrench}.
In this part as well, we omit the unit GW (gigawatt) of the observations and predictions of the electricity
consumption, as well as the one of their corresponding $\rmse$.

Note that this time we do not split anymore the data set into subsets by the half-hours; this is explained in detail below and
comes from two facts: the data set is smaller (and thus the data subsets would be too small) and
we need to abide by an operational constraint as far as the forecasting in France is concerned.

\begin{table}[p]
\caption{\label{tab:charFrench} Some characteristics of the observations $y_t$ (half-hourly mean consumptions)
of the French data set of operational forecasting.}
{\small
\begin{center}
\begin{tabular}{lc}
\hline
Number of days $D$ & 320 \\
Time intervals & Every 30 minutes \\
\hline
Time instances $T$ & 15\,360 ($= 320 \times 48$) \\
Number of experts $N$ & 24 ($=15 + 8 + 1$) \\
\hline
Unit & GW \\
Median of the $y_t$ & 56.33 \\
Bound $B$ on the $y_t$ & 92.76 \\
\hline
\end{tabular}
\end{center}
}
\end{table}

\begin{figure}[p]
\begin{tabular}{cc}
\includegraphics[width=5.5cm]{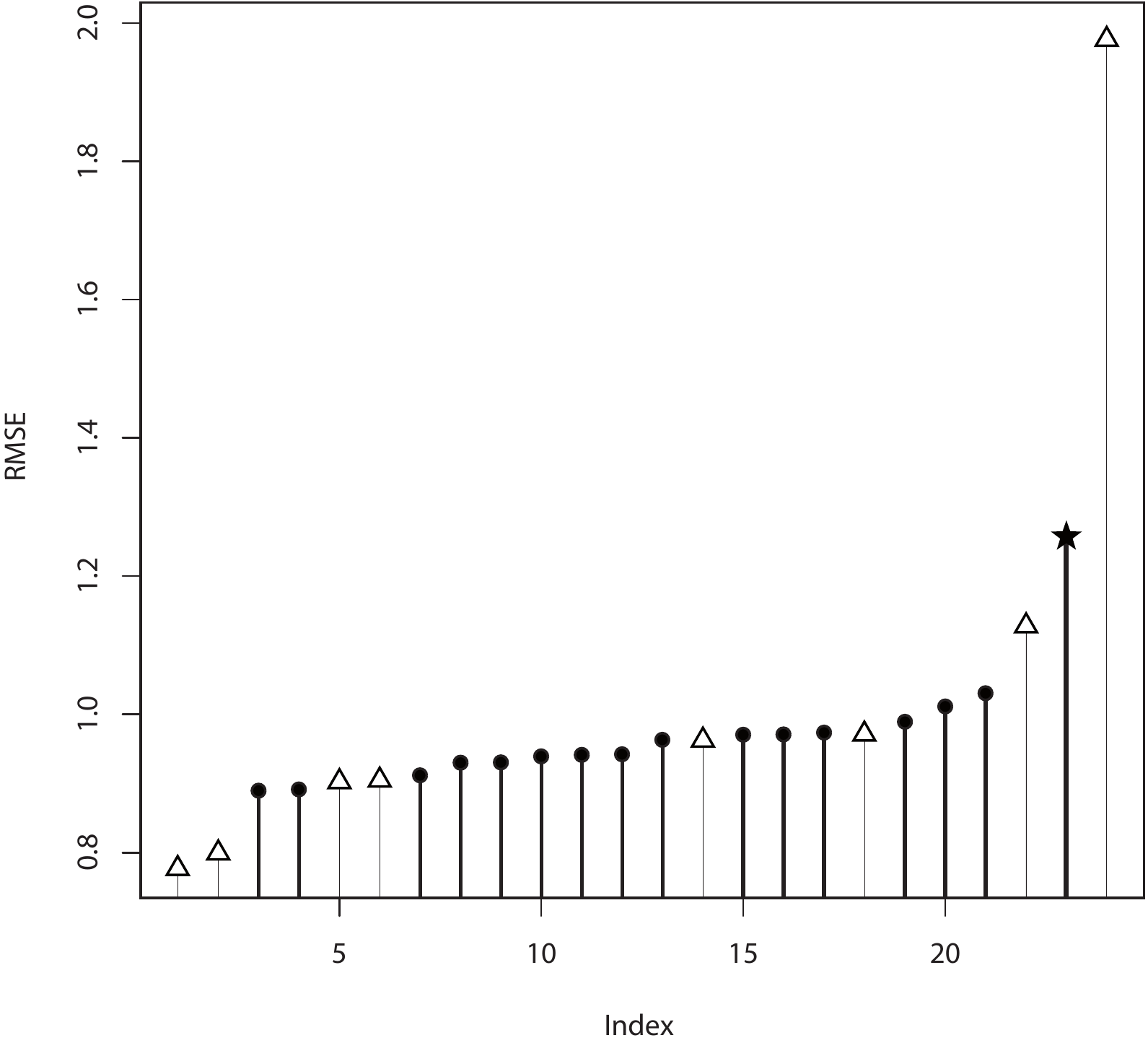}
&
\includegraphics[width=5.5cm]{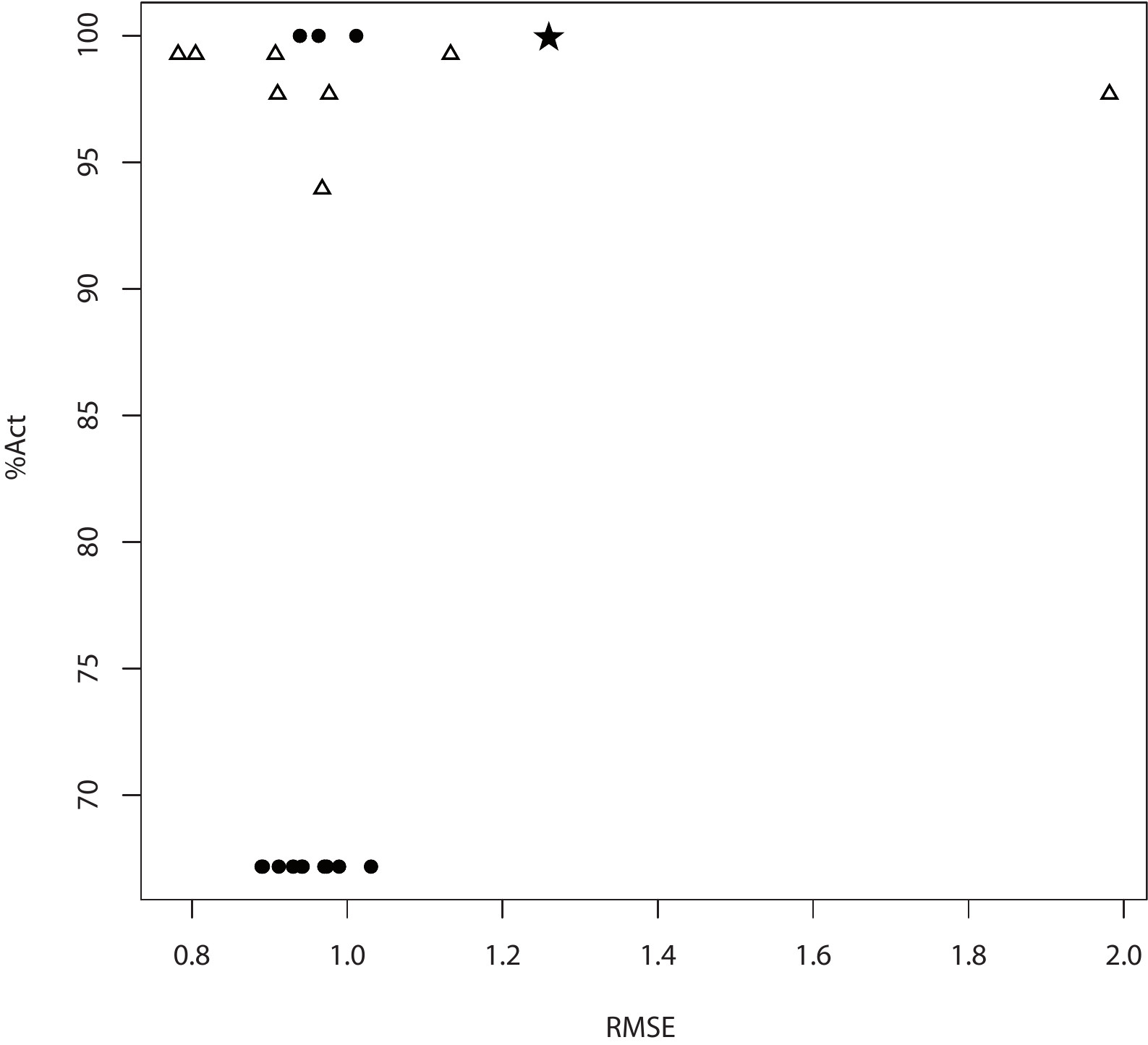}
\end{tabular}
\caption{\label{fig:charFrench} Graphical representations of the performance of the experts
of the French data set: sorted $\rmse$ (left) and
$\rmse$--frequency of activity pairs (right); Eventail experts are depicted by the symbols
$\bullet$, GAM experts are represented by ${\small \triangle}$, while $\star$ stands for the
similarity expert.}
\end{figure}

\begin{table}[p]
\begin{center}
\caption{\label{tab:oraclesFrench} Definition and performance of several (possibly off-line) benchmark
procedures on the French data set; they serve as comparison points for on-line procedures.}
{\small
\begin{tabular}{lcr}
\hline
Name of the benchmark procedure & Formula & Value \\
% \hline
% & & \\
\hline
Uniform sequential aggregation rule & $\rmse(\cU)$ \vspace{.1cm} & $= 0.724$ \\
Uniform convex weight vector & $\displaystyle{\rmse \bigl( (1/24,\ldots,1/24) \bigr)}$ \vspace{.1cm} & $= 0.748$ \\
Best single expert & $\displaystyle{\min_{j = 1,\ldots,24} \ \rmse(j)}$ & $= 0.782$ \\
Best convex weight vector & $\displaystyle{\min_{\bq \in \cX} \ \rmse(\bq)}$ & $= 0.683$ \\
\hline
% & & \\
% \hline
Best compound expert & & \vspace{.15cm}\\
Size at most $m = 50$ & $\displaystyle{\min_{j_1^T \in \cL_{50}} \ \rmse \bigl( j_1^T \bigr)}$ & $= 0.534$ \\
Size at most $m = 100$ & $\displaystyle{\min_{j_1^T \in \cL_{100}} \ \rmse \bigl( j_1^T \bigr)}$ & $= 0.474$ \\
Size at most $m = T - 1 = 15\,359$ & $\displaystyle{\min_{j_1^T \in E_1 \times E_2 \times \ldots \times E_T}
\ \rmse \bigl( j_1^T \bigr)}$ & $= 0.223$ \\
\hline
\end{tabular}
}
\end{center}
\end{table}

\subsection{Brief description of the construction of the considered experts}

The experts we consider here come from three main categories of statistical models:
parametric, semi-parametric, and non-parametric models.
We do so to get experts that are heterogenous and exhibit varied enough behaviors.

The parametric model used to generate the first group of 15 experts is described in~\cite{Bru05} and
is implemented in an EDF software called ``Eventail.'' (For conciseness we refer to them as the Eventail experts.)
This model is based on a nonlinear regression approach that consists of decomposing the electricity load into
a main component including all the seasonality effects of the process together with
a weather-dependant component. To this nonlinear regression model is added an autoregressive correction of the error
of the short-term forecasts of the last seven days. Changing the parameters (the gradient of the temperature, the short-term
correction) of this model led to the indicated 15 experts.

The second group of $8$ experts comes from a generalized additive model
presented in~\cite{Pie09,Pie11} and implemented in the software {\tt R} by the {\tt mgcv} package developed by~\cite{Woo06}.
(We refer to them as the GAM experts.)
The considered generalized additive model imports the idea of the parametric modeling presented above
into a semi-parametric modeling. One of its key advantages is its ability to adapt to changes in consumption habits
while parametric models like Eventail need some a priori knowledge on customers behaviors.
Here again, we derived the $8$ GAM experts
by changing the trend extrapolation effect (which accounts for the yearly economic growth)
or the short-term effects like the one-day-lag effect; these changes affect the reactivity to changes along the run.

The last expert is drastically different from the two previous groups of experts as it relies on a univariate method
(i.e., a method not requiring any exogenous factor like weather conditions); this method is presented in~\cite{Ant06,AnBrCuPo10}.
Its key idea is to assume that the load is driven by an underlying stochastic curve and to view each day as a discrete
recording of this functional process. Forecasts are then performed according to a similarity measure between days.
We call this expert the similarity expert.

\subsection{Benchmark values: performance of the experts and of some oracles}

The characteristics of the experts presented above are depicted in Figure~\ref{fig:charFrench},
here again with a bar plot representing the (sorted) values of the $\rmse$ of the 24 available experts
and a scatter plot relating the $\rmse$ of each of the expert to its frequency of activity.
Out of the 15 Eventail experts, 3 are active all the time; they correspond to the operational model actually used at the R{\&}D center of EDF
and to two variants of it based on different short-term corrections. The other 12 Eventail experts are inactive during the summer
as their predictions are redundant with the 3 main Eventail experts
(they were obtained by changing the gradient of the temperature for the heating part of the load consumption,
which generates differences to the operational model in winter only).
GAM experts are active on an overwhelming fraction of the time and are sleeping only during periods when R{\&}D practitioners
know beforehand that they will perform poorly (e.g., in time periods close to public holidays); the lengths of these periods depend on the parameters
of the expert. Finally, the similarity expert is always active.
\medskip

We report in Table~\ref{tab:oraclesFrench} the performance obtained by most of the
oracles already discussed in Section~\ref{sec:benchm-Slov}. We do not report here
the performance obtained by considering partitions of the time in terms of
the values of the active sets $E_t$, as, on the one hand,
the study of Section~\ref{sec:benchm-Slov} showed that even when the number
of elements $K$ in the partition was large, the compound experts had better
performance, and on the other hand, as the value of $K$ is small here ($K = 7$);
these two facts explain that the performance of the
oracles based on partitions is to expected to be poor on this data set.

We note the disappointing performance of the best single expert with respect to the naive
rule $\cU$. Unlike in Section~\ref{sec:benchm-Slov},
this comes from our experts being more active in challenging situations.
Indeed, the rule $\cU$ also performs better than
the uniform convex weight vector, which induces at each time
instance the same forecast as the rule $\cU$ but for which the loss incurred at
a given time instance is more weighted as more experts are active. All in all,
the poor performance of the best single expert or of the uniform convex weight vector
are caused by the considered specialized experts being more active and more helpful when needed.

From Table~\ref{tab:oraclesFrench} we mostly conclude the following.
The true benchmark values from the first part of the table are
the {\rmse} of the rule $\cU$ --that all fancy rules have to outperform to be
considered worth the trouble-- and the {\rmse} of the best convex weight vector.
The second part of the table indicates that important gains in accuracy
are obtained with compound experts (and therefore, fixed-share type rules
are expected to perform well, which will turn out to be the case).

\subsection{Extension of the considered rules to the operational forecasting constraint}
\label{sec:extopfor}

We consider prediction with an operational constraint required by EDF consisting
of producing half-hourly forecasts every day at 12:00 for the next $24$ hours;
that is, of forecasting simultaneously the next 48 time instances. (The experts
presented above also abide by this constraint.)
The high-level idea is to run the original rules on the data
(called below the base rules),
access to the proposed convex weight vectors only at time
instances of the form $t_k = 48 k + 1$,
and use these vectors for the next 48 time instances, by adapting them
via a renormalization or a share update
to the values of the active sets $E_{t_k+1}, \ldots, E_{t_k+48}$.

We also propose another extension related to the structure of the
set of experts. The latter are of three different types and experts of the same type
are obtained as variants of a given prediction method (GAM, Eventail,
or functional similarity estimation). It would be fair to allocate an initial
weight of $1/3$ to the group of GAM experts, which turns into an initial
weight of $1/24$ to each of the 8 GAM experts; a weight of $1/3$ to
the group formed by the 15 Eventail experts, that is, an initial weight of $1/45$
to each of them; and an initial weight of $1/3$ to the similarity expert.
We denote by $p_{j,0}$ the initial weight of an expert $j$.
We will call fair initial weights the convex weight vector described
above (with components equal to $1/3$, $1/24$, or $1/45$)
and uniform initial weights the vector defined by $p_{j,0} = 1/24$
for all experts $j$. The effect of this on the regret bounds, e.g.,
(\ref{eq:regretboundEWA}) or (\ref{eq:regretboundFSEWA}),
is the replacement of $\ln N$ by $\max_j \, \ln 1/p_{j,0}$.
This does not change the order of magnitude in $T$ of the regret bounds
but only increases them by a multiplicative factor.

All in all,
we denote by $\cW_\eta$ and $\cW_\eta^{\grad}$
the adaptations to the operational constraint of the rules
$\cE_\eta$ and $\cE_\eta^{\grad}$ of Sections~\ref{sec:defEWA} and~\ref{sec:subgrad};
by $\cT_\eta$ and $\cT_\eta^{\grad}$ the ones of the rules $\cS_\eta$ and $\cS_\eta^{\grad}$ of Sections~\ref{sec:defEWA} and~\ref{sec:subgrad};
and by $\cG_{\eta,\alpha}$ and $\cG_{\eta,\alpha}^{\grad}$
the ones of the rules $\cF_{\eta,\alpha}$ and $\cF_{\eta,\alpha}^{\grad}$
described in Section~\ref{sec:defFS}.
For instance, $\cW_\eta$ uses, at time $t = 1,2,\ldots,T$, the weight vector $\bp_t$
defined by
\begin{equation}
\label{eq:defEWAext}
p_{j,t} = \frac{p_{j,0} \,\, e^{\eta R_{48 \lfloor (t-1)/48 \rfloor}( \cE_\eta, j)} \, \ind_{ \{ j \in E_t \} }}{\sum_{k \in E_t}
p_{k,0} \,\, e^{\eta R_{48 \lfloor t/48 \rfloor}( \cE_\eta, k)}}\,,
\end{equation}
for all experts $j$, with the usual convention that empty sums equal 0.
(The notation $\lfloor x \rfloor$ denotes the lower integer part of a real number $x$.)

Similarly, as is illustrated in its statement in Figure~\ref{fig:FSruleExt},
$\cG_{\eta,\alpha}$ basically needs to run an instance of
$\cF_{\eta,\alpha}$ and to access to its proposed weight vector every
48 rounds. Between two such synchronizations,
only share updates (and no loss update) are performed,
to deal with the fact that experts are specialized.
Indeed, the values of the sets of active experts $E_t$
may (and do) vary within a one-day-ahead period of time.
\begin{figure}[t]
\begin{minipage}[l]{\textwidth}
\emph{Parameters}: $\eta >0$ and $0\leq \alpha\leq 1$, as well as an initial convex weight vector
$\bigl( p_{1,0} , \, \ldots, \, p_{N,0} \bigr)$
\smallskip\noindent

\emph{Initialization}: $(w_{1,0},\ldots,w_{N,0}) = \bigl( p_{1,0} \, \ind_{ \{ 1 \in E_1 \} }, \, \ldots,
\, p_{N,0} \, \ind_{ \{ N \in E_1 \} } \bigr)$
\smallskip\noindent

{\em For} each round $t = 1,2,\ldots,T$,
  \smallskip\noindent
  \begin{itemize}
  \item[(1)] $\displaystyle{\wh{y}_t = \frac{1}{\sum_{k=1}^N w_{k,t-1}} \sum_{j=1}^N w_{j,t-1} \, f_{j,t}}$\,; \vspace{.05cm} \\
  \item[(2)] [loss and share updates] \\ if $t = 48 k$ for some $k$,
  observe $y_{t-47},\ldots,y_t$ and take\footnote{$\bp_{t+1}(\cF_{\eta,\alpha})$ is
  the convex weight vector chosen by the rule $\cF_{\eta,\alpha}$ after seeing the sequence
  of observations $y_1,\ldots,y_t$ and the corresponding expert predictions; we use here the same
  notation as in Section~\ref{sec:calibr}, where we indicated in parentheses the name of the rule
  whenever it was needed. Here, the rule $\cG_{\eta,\alpha}$
  thus synchronizes again with $\cF_{\eta,\alpha}$ at steps $t$ of the form
  $t_k = 48k$ for some $k$.} $(w_{1,t},\ldots,w_{N,t}) = \bp_{t+1}(\cF_{\eta,\alpha})$; \vspace{.05cm} \\
  \item[(3)] [share update] \\ otherwise (when $t$ is not a multiple of 48),
  let $w_{j,t} = 0$ if $j \not\in E_{t+1}$ and
   \[
w_{j,t} =\frac{1}{\abs[E_{t+1}]}\sum_{i \in E_t\setminus E_{t+1}}\!\!\!w_{i,t-1}+ \frac{\alpha}{\abs[E_{t+1}]}\sum_{i \in E_t\cap E_{t+1}}\!\!\!w_{i,t-1}+(1-\alpha)\,\ind_{ \{ j\in E_t\cap E_{t+1} \} }\, w_{j,t-1}
   \]
   if $j \in E_{t+1}$ (with the convention that an empty sum is null).
  \end{itemize}
\end{minipage}
\rule{\linewidth}{.5pt}
\caption{\label{fig:FSruleExt} The extension $\cG_{\eta,\alpha}$
of the (basic) fixed-share aggregation rule $\cF_{\eta,\alpha}$ to
operational forecasting.}
\end{figure}

Theoretical bounds on the regret can be proved
since, as is clear from the algorithmic statements of the extensions,
the weights output by the base rules are, for all $t$,
close to the ones of their adaptations (and of course,
coincide with them at the time instances $t_k$). This is because
these weights are computed on almost the same sets of losses;
these sets differ by at most 47 losses, the ones between
the last $t_k$ and the current instance $t$.
A quantification of this fact and a sketch of a regret bound, e.g.,
for $\cW_\eta$, are provided in the appendix (Section~\ref{sec:operadapt}).

\subsection{Results obtained with constant values of the parameters}
\label{sec:grid3}

The performance of the extensions $\cW_\eta$, $\cW_\eta^{\grad}$, $\cT_\eta$, and $\cT_\eta^{\grad}$ described above is
summarized in Table~\ref{tab:offlineFrench1}.
We note that the gradient versions of the forecasters (for both priors)
outperform the comparison point formed by the {\rmse} of the best convex weight vector, equal to $0.696$,
and which was the only interesting benchmark value among the oracles of the first part of
Table~\ref{tab:oraclesFrench}.
They do so by a relative factor of about $5\,\%$; on the other hand,
their basic versions (in case of a fair prior) get only a slightly improved performance
with respect to this comparison point.
It is also worth noting that the performance
of the gradient versions is not sensitive to the initial allocation of weights.

\begin{remark}
\label{rk:perftheo2}
Here again, as already mentioned for the Slovakian data set in Section~\ref{sec:perfSlovCst},
the best constant choices in hindsight are far away from the theoretically
optimal ones, given by values $\eta^\star$ of the order of $10^{-6}$ on the present data set.
For such small values of $\eta$, the rules are basically equivalent to the uniform
aggregation rule $\cU$, as is indicated by the performance reported in Table~\ref{tab:offlineFrench1}.
\end{remark}

The performance of the extensions $\cG_{\eta,\alpha}$ and $\cG_{\eta,\alpha}^{\grad}$
described above is summarized in Table~\ref{tab:offlineFrench2}.
(It turned out that the performance of the algorithms did not
depend much on whether the initial weight allocation was fair or uniform and
we report only the results obtained by the latter in the sequel.)
The comparison points are given by the best compound experts studied in
Table~\ref{tab:oraclesFrench}, which exhibited an excellent performance. This is why
we expected and actually see a significant gain of performance for the aggregation rules when resorting to
forecasters tracking the performance of the compound experts.
Table~\ref{tab:offlineFrench2} shows a relative improvement in the performance of about $5\,\%$
with respect to the results of Table~\ref{tab:offlineFrench1}.

\subsection{Results obtained with a fully online tuning of the parameters}
\label{sec:fulladapt}

\begin{table}[p]
\caption{\label{tab:offlineFrench1}
Performance obtained by the sequential aggregation rules
$\cW_\eta$, $\cW_\eta^{\grad}$, $\cT_\eta$, and $\cT_\eta^{\grad}$ for various choices of
$\eta$; the smallest {\rmse} obtained for each rule is underlined.}
{\small
\begin{center}
\begin{tabular}{lll*{7}{c}}
\hline
Values & of $\eta$ &  Prior & $10^{-6}$ & $10^{-5}$  & $2 \times 10^{-4}$ & $10^{-3}$ & $2 \times 10^{-2}$ & $10^{-1}$ & $2$ \\
\hline
{\rmse} & $\mathcal{W}_\eta$ & (unf.) & $0.724$ & $0.722$ & $\underline{0.718}$ & $0.731$ & $0.784$ & $0.783$ & $0.784$\\
 	    & $\mathcal{W}_\eta$ & (fair) & $0.736$ & $0.731$ & \underline{$0.684$} & $0.722$ & $0.785$ & $0.784$ & $0.785$ \\
	    & $\mathcal{W}_\eta^{\grad}$ & (unf.) & $0.724$ & $0.722$ & $0.705$ & $0.683$ & $0.631$ & $0.640$ & \underline{$0.629$} \\
	    & $\mathcal{W}_\eta^{\grad}$ & (fair) & $0.737$ & $0.733$ & $0.697$ & $0.674$ & \underline{$0.633$} & $0.641$ & $0.640$ \\
	    & $\mathcal{T}_\eta$ & (unf.) & $0.724$ & $0.722$ & $\underline{0.718}$ & $0.731$ & $0.785$ & $0.783$ & $0.752$ \\
 	    & $\mathcal{T}_\eta$ & (fair) & $0.736$ & $0.731$ & \underline{$0.684$} & $0.721$ & $0.786$ & $0.784$  & $0.753$\\
	    & $\mathcal{T}_\eta^{\grad}$ & (unf.) & $0.724$  & $0.712$ & $0.705$ & $0.683$ & $\underline{0.631}$ & $0.640$ & $0.741$ \\
	    & $\mathcal{T}_\eta^{\grad}$ & (fair) & $0.737$ & $0.733$ & $0.697$ & $0.674$  & \underline{$0.633$} & $0.641$ & $0.855$\\
\hline
\end{tabular}
%\begin{tabular}{lll*{7}{c}}
%\hline
%Values \hspace{-.2cm} & of $\eta$ & Prior & $10^{-6}$ & $10^{-5}$ & $10^{-4}$ & $2 \times 10^{-4}$ & $10^{-3}$ & $5 \times 10^{-3}$ & $10^{-2}$ \\
%\hline
%{\rmse} \hspace{-.2cm} & $\cW_\eta$ & (unf.) & 0.724 & 0.722 & \underline{0.718} & & 0.731 & & 0.788 \\
%\hspace{-.2cm} & $\cW_\eta$ & (fair) & 0.736 & 0.731 & 0.695 & \underline{0.683} & 0.722 & & 0.789 \\
%\hspace{-.2cm} & $\cW_\eta^{\grad}$ & (unf.) & 0.724 & 0.722 & 0.712 & & 0.683 & \underline{0.650} & 0.668 \\
%\hspace{-.2cm} & $\cW_\eta^{\grad}$ & (fair) & 0.737 & 0.733 & 0.711 & & 0.674 & \underline{0.651} & 0.670 \\
%\hspace{-.2cm} & $\cT_\eta$ & (...) & \\
%\hspace{-.2cm} & $\cT_\eta$ & (...) & \\
%\hspace{-.2cm} & $\cT^{\grad}_\eta$ & (...) & \\
%\hspace{-.2cm} & $\cT^{\grad}_\eta$ & (...) & \\
%\hline
%\end{tabular}
\end{center}
}
\end{table}

\begin{table}[p]
\caption{\label{tab:offlineFrench2}
Performance obtained by the sequential aggregation rules
$\cG_{\eta,\alpha}$ and $\cG_{\eta,\alpha}^{\grad}$
run with an initial uniform allocation of the weights
for various choices of $\eta$ and $\alpha$;
the smallest {\rmse} obtained for each rule is underlined.}
{\small
\begin{center}
\begin{tabular}{ll*{9}{c}}
\hline
Values & of $\eta$ & 0.01 & 0.01 & 0.01 & 1 & 1 & 1 & 500 & 500 & 500 \\
       & of $\alpha$ & 0.001 & 0.01 & 0.05 & 0.001 & 0.01 & 0.05 & 0.001 & 0.01 & 0.05 \\
\hline
{\rmse} & $\cG_{\eta,\alpha}$ & 0.678 & 0.683 & 0.704 & 0.711 & 0.659 & 0.652 & 0.674 & 0.633 & \underline{0.632} \\
        & $\cG_{\eta,\alpha}^{\grad}$ & 0.646 & 0.669 & 0.700 & 0.622 & \underline{0.598} & 0.637 & 0.683 & 0.675 & 0.671 \\
\hline
\end{tabular}
\end{center}
}
\end{table}

\begin{table}[p]
\enlargethispage{5cm}
\caption{\label{tab:onlineFrench1}
Performance obtained by
the rules $\cW_\eta$,  $\cW_\eta^{\grad}$, $\cT_\eta$, and $\cT_\eta^{\grad}$
for the best constant choice of $\eta$ in hindsight
and when used as keystones of a meta-rule selecting
sequentially the values of $\eta$ based on an adaptive grid;
results are reported for both the uniform and fair priors.}
{\small
\begin{center}
\begin{tabular}{ll*{4}{c}}
\hline
& & \multicolumn{2}{c}{Uniform prior} &  \multicolumn{2}{c}{Fair prior} \\
& & Best constant $\eta$ & Adaptive grid & Best constant $\eta$ & Adaptive grid \\
\hline
{\rmse} of & $\cW_\eta$ & 0.718 & 0.724 & 0.684 & 0.696 \\
           & $\cW_\eta^{\grad}$ & 0.629 & 0.640 & 0.633 & 0.644 \\
           & $\mathcal{T}_\eta$ & 0.718 & 0.723 & 0.684 & 0.698 \\
           & $\mathcal{T}_\eta^{\grad}$ & 0.631 & 0.640 & 0.633 & 0.645 \\
%           & $\cT_\eta$ & (...) & ... & ... \\
%           & $\cT_\eta$ & (...) & ... & ... \\
%           & $\cT^{\grad}_\eta$ & (...) & ... & ... \\
%           & $\cT^{\grad}_\eta$ & (...) & ... & ... \\
\hline
\end{tabular}
\end{center}
}
\end{table}

\begin{table}[p]
\caption{\label{tab:onlineFrench2}
Performance obtained by
the rules $\cG_{\eta,\alpha}$ and $\cG_{\eta,\alpha}^{\grad}$
run with an initial uniform weight allocation
for the best constant choices of $\eta$ and $\alpha$ in hindsight (left)
and when used as keystones of a meta-rule selecting
sequentially the values of $\eta$ based on an adaptive grid
and the values of $\alpha$ according to a fixed grid (right).}
{\small
\begin{center}
\begin{tabular}{llcc}
\hline
& & Best constant pair $(\eta,\alpha)$ & Adaptive grid \\
\hline
{\rmse} of & $\cG_{\eta,\alpha}$ & 0.632 & 0.658 \\
           & $\cG_{\eta,\alpha}^{\grad}$ & 0.598 & 0.623 \\
\hline
\end{tabular}
\end{center}
}
\end{table}

In Sections~\ref{sec:calibr} and~\ref{sec:onlc-Slov}
we indicated that our simulations showed that the step of the grid was not
too crucial parameter and that the results were not too sensitive to it;
we however did not clarify how to choose the maximal (and also the minimal) possible value(s) of $\eta$ in the considered grids,
i.e., how to determine the right scaling for $\eta$.
The procedure is based on the observation that in Figures~\ref{fig:etaonline} and~\ref{fig:etaalphaonline} of Section~\ref{sec:onlc-Slov}
the selected parameters $\wh{\eta}_t$ are eventually constant or vary in a small range. It thus simply suffices to ensure
that the constructed grid covers a large enough span.
This can be implemented by extending online the considered grid as follows.
We let the user fix an arbitrary finite starting grid, say, reduced to $\{1\}$.
At any time $t$ when the selected parameter $\wh{\eta}_{t-1}$ is an endpoint of the grid,
we enlarge it by adding the values $2^r \, \wh{\eta}_{t-1}$, for $r \in \{ 1,2,3 \}$,
respectively, for $r \in \{ -1,-2,-3 \}$, if the endpoint was the upper limit,
respectively, the lower limit of the grid.
(We tested different factors than the factor of 2 considered here and also tried to increase the grid
with more than three points; no such change had an important impact on the performance.)
The possible choices for $\alpha$ are in the (known) bounded range $[0,1]$ and therefore
no scaling issue takes place. We considered a fixed grid of possible $\alpha$
given by
\[
\alpha \in \bigl\{ 0, \,\, 0.005, \,\, 0.01, \,\, 0.05, \,\, 0.1, \,\, 0.2, \,\, 0.5, \,\, 1 \bigr\}\,.
\]

The performance of this adaptive construction of the grids used by the meta-rules
with respect to the best constant choices in hindsight
is summarized in Tables~\ref{tab:onlineFrench1} and~\ref{tab:onlineFrench2}.
We observe that the now fully sequential character of the meta-rule comes at a limited cost in the performance.
(That cost would be almost insignificant if a training period was allowed, so as to start the evaluation period
with a grid already large enough.)

\subsection{Robustness study of the considered aggregation rules}
\label{sec:robust}

\begin{figure}[p]
\begin{center}
\includegraphics[width=9cm]{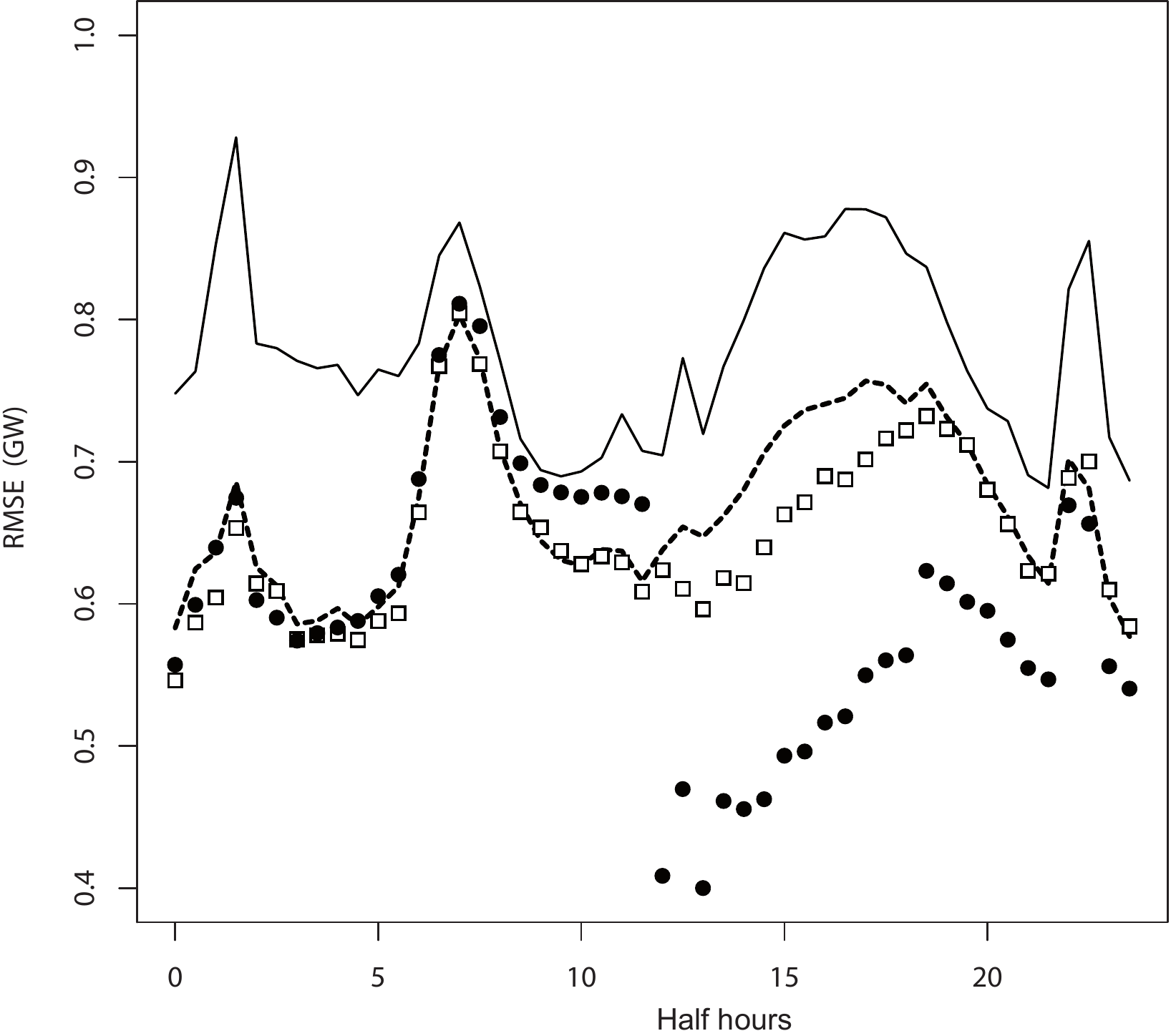}
\end{center}
\caption{\label{fig:Hourly-RMSE}
Half-hourly {\rmse} of the
meta-rules based on the rules
$\cW_\eta^{\grad}$ (symbol: $\square$) and $\cG_{\eta,\alpha}^{\grad}$
(symbol: $\bullet$);
as well as the ones of the best overall single expert (solide line)
and of the best overall convex weight vector (dashed line).
}
\end{figure}

\begin{figure}[p]
\begin{center}
\includegraphics[width=9cm]{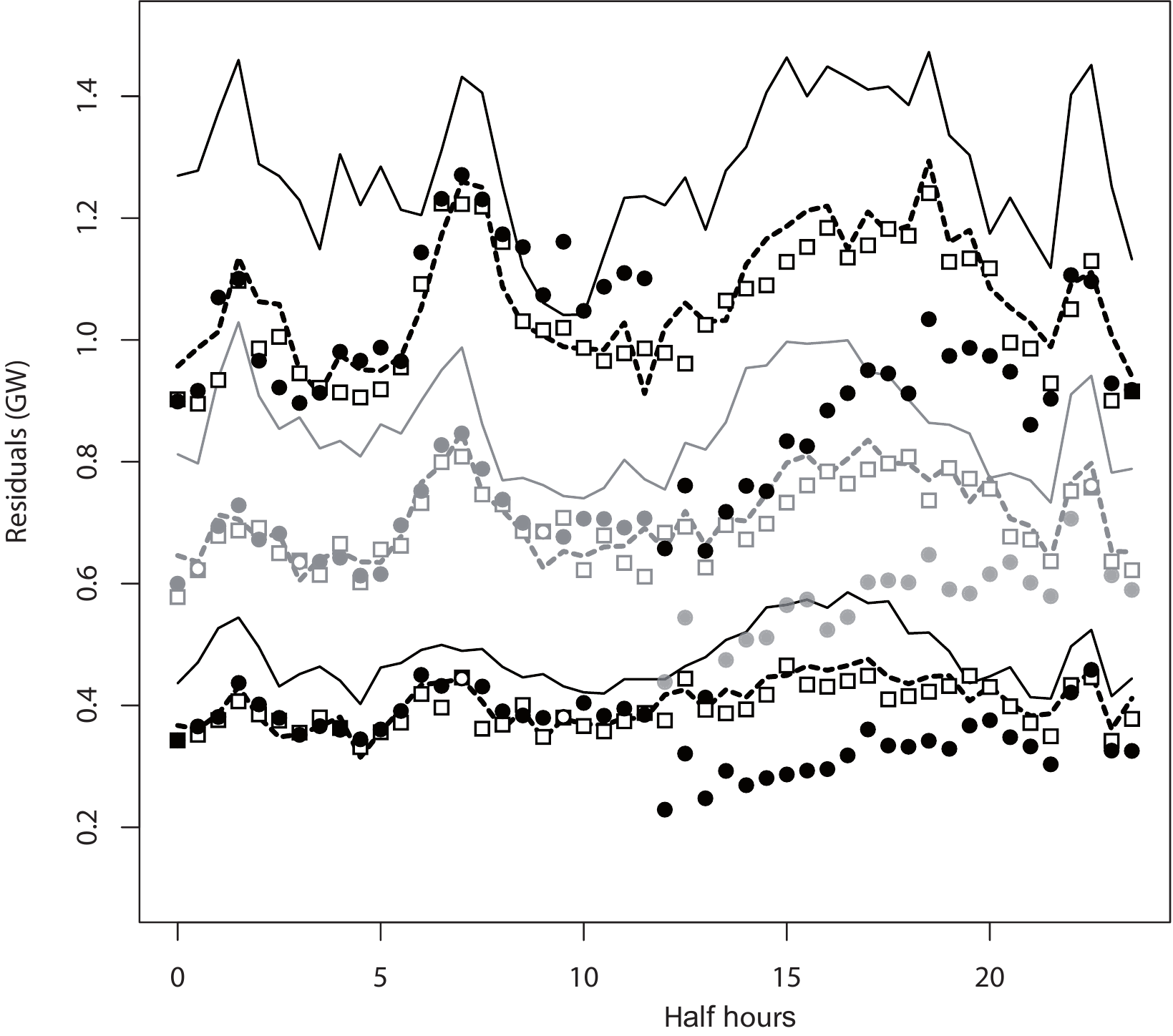}
\end{center}
\caption{\label{fig:IQS} Using the same rules and benchmarks as in Figure~\ref{fig:Hourly-RMSE},
with the same legend: $50\,\%$ (black), $75\,\%$ (grey), and $90\,\%$ (black) quantiles of the absolute values of
the residuals, grouped per half hours.
}
\end{figure}

In this section we move from the study of global average behaviors of the aggregation
rules (as measured by their {\rmse}) to a more individual analysis, based on the scattering of the
prediction residuals $\wh{y}_t - y_t$.
The {\rmse} is indeed a global criterion
and we want to check that the overall good performance does not come
at the cost of local disasters in the accuracy of the aggregated forecasts.
To that end we split the data set by the half hours into $48$ sub-data sets;
for each of these subsets we compute the {\rmse}s of some of the benchmarks and aggregation rules discussed above and
study also the scattering of the (absolute values of the) prediction residuals.
To do so we consider two fully sequential
aggregation rules, namely, the meta-rules based on
families of $\cW_\eta^{\grad}$ and $\cG_{\eta,\alpha}^{\grad}$ run with initial uniform weight allocations.
We use as benchmarks the (overall) best single expert and the (overall) best convex weight vector, whose
performance was reported in Table~\ref{tab:oraclesFrench}.

Figure~\ref{fig:Hourly-RMSE} plots the half-hourly {\rmse} of these two aggregation rules and of these two
benchmarks. It shows that the performance of the rule based on exponential weighted averages
is, uniformly over the $48$ elements of the partition
of days in half hours, at least as good as the one of the best constant convex combination of the expert forecasts.
The performance of the rule based on fixed-share aggregation rules is intriguing: its accuracy is significantly improved
with respect to the one of the latter benchmark between 12:00 and 21:00 but is also slightly
worse between 6:00 and 12:00.
It thus seems that this rule has excellent performance on very short-term horizon and would probably strongly benefit
from an intermediate update around midnight (this is however not the purpose of the present study: intra-day forecasting
is left for future research).
% We can provide no reason for this behavior yet; a further study of this behavior and its benefits is left to future research.
A similar behavior is observed in Figure~\ref{fig:IQS}, which depicts the
medians, the third quartiles, and the $90\,\%$ quantiles of the absolute values of
the residuals grouped by half hours. In addition,
we see that the distributions of the errors of the aggregation rules are more concentrated than the ones
of the best benchmarks, which indicates that
their good overall performance does not come at the cost of some local disasters in the quality of the predictions.

All in all, we conclude that the best aggregation rules never encounter large prediction errors in comparison to the best expert or
to the best convex combination of experts and often encounter much smaller such errors. This is strongly in favor of
their use in an industrial context where large errors can be highly prejudicious (potential issues
range from financial penalties to black outs).
In a nutshell, aggregation rules are seen to reduce the risk of prediction, which is one important pro for operational forecasting.

\section{Conclusions}
\label{sec:ccl}

On the theoretical side, we reviewed and extended known aggregation rules for the case of specialized (sleeping) experts.
First, we provided a general analysis of the specialist aggregation rules
of~\cite{FrScSiWa97} for all convex loss functions, while the original reference
needed an ad hoc analysis for each loss function of interest. Second, we showed
how the fixed-share rules of~\cite{HeWa98} can accommodate specialized experts: they form a natural
and efficient alternative to the specialist aggregation rules.
Finally, for all these rules, as well as the exponentially weighted average ones,
we indicated how to extend them so as to take into account
some operational constraint of outputting simultaneous forecasts for a fixed number of future
time instances.

We then followed a general methodology to study the performance of these rules on real data
of electricity consumption. In particular, we provided fully adaptive methods that
can tune online their parameters based on adaptive grids; doing so, they outperform clearly
the rules tuned with the theoretically optimal parameters.
All in all, for the two data sets at hand the best rules, given by fixed-share type rules,
improve on the accuracy of the best constant convex combination of the experts
by about $5\,\%$ (Slovakian data set) to about $15\,\%$ (French data set).
In addition, we noted that resorting to the gradient trick described in Section~\ref{sec:subgrad}
always improved the performance of the underlying aggregation rule.
Finally, the raw improvement in terms of the global performance, as measured by the {\rmse},
of the sequential aggregation rules over the (convex combinations of) experts,
also comes together with a reduction of the risk of large errors: the
studied aggregation rules are more robust than the base forecasters they are using.

\begin{acknowledgements}
We thank the anonymous reviewers and associated editor for their valuable comments
and feedback, which improved drastically the exposition of our results and conclusions.
Marie Devaine and Pierre Gaillard carried out this research
while completing internships at EDF R{\&}D, Clamart;
this article is based on the technical reports (\citep{DeGoSt09,Pierre}) written therefor.
Gilles Stoltz was partially supported by the French ``Agence Nationale pour la Recherche''
under grant JCJC06-137444 ``From applications to theory in learning and adaptive statistics''
and by the PASCAL Network of Excellence under EC grant {no.} 506778.
\end{acknowledgements}

{\small
\bibliographystyle{plainnat}
\bibliography{biblio-Devaine-Goude-Stoltz-Gaillard}

\begin{thebibliography}{34}
\providecommand{\natexlab}[1]{#1}
\providecommand{\url}[1]{\texttt{#1}}
\expandafter\ifx\csname urlstyle\endcsname\relax
  \providecommand{\doi}[1]{doi: #1}\else
  \providecommand{\doi}{doi: \begingroup \urlstyle{rm}\Url}\fi

\bibitem[Antoniadis et~al.(2006)Antoniadis, Paparoditis, and Sapatinas]{Ant06}
A.~Antoniadis, E.~Paparoditis, and T.~Sapatinas.
\newblock A functional wavelet--kernel approach for time series prediction.
\newblock \emph{Journal of the Royal Statistical Society: Series B},
  68\penalty0 (5):\penalty0 837--857, 2006.

\bibitem[Antoniadis et~al.(2010)Antoniadis, Brossat, Cugliari, and
  Poggi]{AnBrCuPo10}
A.~Antoniadis, X.~Brossat, J.~Cugliari, and J.M. Poggi.
\newblock Clustering functional data using wavelets.
\newblock In \emph{Proceedings of the Nineteenth International Conference on
  Computational Statistics (COMPSTAT)}, 2010.

\bibitem[Auer et~al.(2002)Auer, Cesa-Bianchi, and Gentile]{AuCeGe02}
P.~Auer, N.~Cesa-Bianchi, and C.~Gentile.
\newblock Adaptive and self-confident on-line learning algorithms.
\newblock \emph{Journal of Computer and System Sciences}, 64:\penalty0 48--75,
  2002.

\bibitem[Blum(1997)]{Bl97}
A.~Blum.
\newblock Empirical support for winnow and weighted-majority algorithms:
  {R}esults on a calendar scheduling domain.
\newblock \emph{Machine Learning}, 26:\penalty0 5--23, 1997.

\bibitem[Blum and Mansour(2007)]{BlMa07}
A.~Blum and Y.~Mansour.
\newblock From external to internal regret.
\newblock \emph{Journal of Machine Learning Research}, 8:\penalty0 1307--1324,
  2007.

\bibitem[Borodin et~al.(2000)Borodin, El-Yaniv, and Gogan]{Gog00}
A.~Borodin, R.~El-Yaniv, and V.~Gogan.
\newblock On the competitive theory and practice of portfolio selection.
\newblock In \emph{Proceedings of the Fourth Latin American Symposium on
  Theoretical Informatics (LATIN'00)}, pages 173--196, 2000.

\bibitem[Bruhns et~al.(2005)Bruhns, Deurveilher, and Roy]{Bru05}
A.~Bruhns, G.~Deurveilher, and J.-S. Roy.
\newblock A non-linear regression model for mid-term load forecasting and
  improvements in seasonnality.
\newblock In \emph{Proceedings of the Fifteenth Power Systems Computation
  Conference (PSCC)}, 2005.

\bibitem[Bunn and Farmer(1985)]{Bu85}
D.~W. Bunn and E.~D. Farmer.
\newblock \emph{Comparative Models for Electrical Load Forecasting}.
\newblock John Wiley and Sons Inc., New York, 1985.

\bibitem[Cesa-Bianchi and Lugosi(2003)]{CeLu03}
N.~Cesa-Bianchi and G.~Lugosi.
\newblock Potential-based algorithms in on-line prediction and game theory.
\newblock \emph{Machine Learning}, 51:\penalty0 239--261, 2003.

\bibitem[Cesa-Bianchi and Lugosi(2006)]{CeLu06}
N.~Cesa-Bianchi and G.~Lugosi.
\newblock \emph{Prediction, Learning, and Games}.
\newblock Cambridge University Press, 2006.

\bibitem[Cesa-Bianchi et~al.(2007)Cesa-Bianchi, Mansour, and Stoltz]{CeMaSt07}
N.~Cesa-Bianchi, Y.~Mansour, and G.~Stoltz.
\newblock Improved second-order inequalities for prediction under expert
  advice.
\newblock \emph{Machine Learning}, 66:\penalty0 321--352, 2007.

\bibitem[Cover(1991)]{Cov91}
T.M. Cover.
\newblock Universal portfolios.
\newblock \emph{Mathematical Finance}, 1:\penalty0 1--29, 1991.

\bibitem[Dani et~al.(2006)Dani, Madani, Pennock, Sanghai, and
  Galebach]{DaMaPeSaGa06}
V.~Dani, O.~Madani, D.~Pennock, S.~Sanghai, and B.~Galebach.
\newblock An empirical comparison of algorithms for aggregating expert
  predictions.
\newblock In \emph{Proceedings of the Twenty-Second Conference on Uncertainty
  in Artificial Intelligence (UAI)}, 2006.

\bibitem[Dashevskiy and Luo(2011)]{Network}
M.~Dashevskiy and Z.~Luo.
\newblock Time series prediction with performance guarantee.
\newblock \emph{{IET} Communications}, 5:\penalty0 1044--1051, 2011.

\bibitem[de~Rooij and van Erven(2009)]{Dutch}
S.~de~Rooij and T.~van Erven.
\newblock Learning the switching rate by discretising {B}ernoulli sources
  online.
\newblock In \emph{Proceedings of the Twelfth International Conference on
  Artificial Intelligence and Statistics (AISTATS)}, 2009.

\bibitem[Devaine et~al.(2009)Devaine, Goude, and Stoltz]{DeGoSt09}
M.~Devaine, Y.~Goude, and G.~Stoltz.
\newblock Aggregation of sleeping predictors to forecast electricity
  consumption.
\newblock Technical report, {\'E}cole normale sup\'erieure, Paris and EDF R\&D,
  Clamart, July 2009.
\newblock Available at
  \url{http://www.math.ens.fr/\%7Estoltz/DeGoSt-report.pdf}.

\bibitem[Dordonnat et~al.(2008)Dordonnat, Koopman, Ooms, Dessertaine, and
  Collet]{Dor08}
V.~Dordonnat, S.J. Koopman, M.~Ooms, A.~Dessertaine, and J.~Collet.
\newblock An hourly periodic state space model for modelling {F}rench national
  electricity load.
\newblock \emph{International Journal of Forecasting}, 24:\penalty0 566--587,
  2008.

\bibitem[Freund et~al.(1997)Freund, Schapire, Singer, and Warmuth]{FrScSiWa97}
Y.~Freund, R.~Schapire, Y.~Singer, and M.~Warmuth.
\newblock Using and combining predictors that specialize.
\newblock In \emph{Proceedings of the Twenty-Ninth Annual ACM Symposium on the
  Theory of Computing (STOC)}, pages 334--343, 1997.

\bibitem[Gaillard et~al.(2011)Gaillard, Goude, and Stoltz]{Pierre}
P.~Gaillard, Y.~Goude, and G.~Stoltz.
\newblock A further look at the forecasting of the electricity consumption by
  aggregation of specialized experts.
\newblock Technical report, {\'E}cole normale sup\'erieure, Paris and EDF R\&D,
  Clamart, July 2011.
\newblock Updated February 2012; available at
  \url{http://ulminfo.fr/\%7Epgaillar/doc/GaGoSt-report.pdf}.

\bibitem[Gerchinovitz et~al.(2008)Gerchinovitz, Mallet, and Stoltz]{GeMaSt08}
S.~Gerchinovitz, V.~Mallet, and G.~Stoltz.
\newblock A further look at sequential aggregation rules for ozone ensemble
  forecasting.
\newblock Technical report, INRIA Paris-Rocquencourt and {\'E}cole normale
  sup\'erieure, Paris, September 2008.
\newblock Available at
  \url{http://www.math.ens.fr/\%7Estoltz/GeMaSt-report.pdf}.

\bibitem[Goude(2008{\natexlab{a}})]{Gou08}
Y.~Goude.
\newblock \emph{M{\'e}lange de pr{\'e}dicteurs et application {\`a} la
  pr{\'e}vision de consommation {\'e}lectrique}.
\newblock PhD thesis, Universit{\'e} Paris-Sud XI, January 2008{\natexlab{a}}.

\bibitem[Goude(2008{\natexlab{b}})]{Gou08_2}
Y.~Goude.
\newblock Tracking the best predictor with a detection based algorithm.
\newblock In \emph{Proceedings of the Joint Statistical Meetings (JSP)},
  2008{\natexlab{b}}.
\newblock See the section on Statistical Computing.

\bibitem[Herbster and Warmuth(1998)]{HeWa98}
M.~Herbster and M.~Warmuth.
\newblock Tracking the best expert.
\newblock \emph{Machine Learning}, 32:\penalty0 151--178, 1998.

\bibitem[Jacobs(2011)]{Jac11}
A.Z. Jacobs.
\newblock Adapting to non-stationarity with growing predictor ensembles.
\newblock Master's thesis, Northwestern University, 2011.

\bibitem[Kleinberg et~al.(2008)Kleinberg, Niculescu-Mizil, and
  Sharma]{KlNiSh08}
R.D. Kleinberg, A.~Niculescu-Mizil, and Y.~Sharma.
\newblock Regret bounds for sleeping experts and bandits.
\newblock In \emph{Proceedings of the Twenty-First Annual Conference on
  Learning Theory (COLT)}, pages 425--436, 2008.

\bibitem[Mallet(2010)]{Mal10}
V.~Mallet.
\newblock Ensemble forecast of analyses: coupling data assimilation and
  sequential aggregation.
\newblock \emph{Journal of Geophysical Research}, 115\penalty0 (D24303), 2010.

\bibitem[Mallet et~al.(2009)Mallet, Stoltz, and Mauricette]{MaStMa09}
V.~Mallet, G.~Stoltz, and B.~Mauricette.
\newblock Ozone ensemble forecast with machine learning algorithms.
\newblock \emph{Journal of Geophysical Research}, 114\penalty0 (D05307), 2009.

\bibitem[Monteleoni and Jaakkola(2003)]{MoJa03}
C.~Monteleoni and T.~Jaakkola.
\newblock Online learning of non-stationary sequences.
\newblock In \emph{Advances in Neural Information Processing Systems (NIPS)},
  volume~16, pages 1093--1100, 2003.

\bibitem[Monteleoni et~al.(2011)Monteleoni, Schmidt, Saroha, and
  Asplund]{MoCIDU}
C.~Monteleoni, G.~Schmidt, S.~Saroha, and E.~Asplund.
\newblock Tracking climate models.
\newblock \emph{Journal of Statistical Analysis and Data Mining}, 4:\penalty0
  372--392, 2011.
\newblock Special issue ``Best of CIDU 2010''.

\bibitem[Pierrot and Goude(2011)]{Pie11}
A.~Pierrot and Y.~Goude.
\newblock Short-term electricity load forecasting with generalized additive
  models.
\newblock In \emph{Proceedings of the Sixteenth International Conference on
  Intelligent System Application to Power Systems (ISAP)}, 2011.

\bibitem[Pierrot et~al.(2009)Pierrot, Laluque, and Goude]{Pie09}
A.~Pierrot, N.~Laluque, and Y.~Goude.
\newblock Short-term electricity load forecasting with generalized additive
  models.
\newblock In \emph{Proceedings of the Third International Conference on
  Computational and Financial Econometrics (CFE)}, 2009.

\bibitem[Stoltz and Lugosi(2005)]{StLu05}
G.~Stoltz and G.~Lugosi.
\newblock Internal regret in on-line portfolio selection.
\newblock \emph{Machine Learning}, 59:\penalty0 125--159, 2005.

\bibitem[Vovk and Zhdanov(2008)]{VoZh08}
V.~Vovk and F.~Zhdanov.
\newblock Prediction with expert advice for the {B}rier game.
\newblock In \emph{Proceedings of the Twenty-Fifth International Conference on
  Machine Learning (ICML)}, 2008.

\bibitem[Wood(2006)]{Woo06}
S.N. Wood.
\newblock \emph{Generalized Additive Models: An Introduction with R}.
\newblock Chapman and Hall/CRC, 2006.

\end{thebibliography}
}

\appendix

\section{Proof of Theorem~\ref{th:spec}}
\label{sec:app2}

\begin{proof}
One can show by induction that the vectors $\bw_t$ are convex weight vectors.
We use the notation defined in Section~\ref{sec:subgrad} for the normalization $\bq^E$
of convex weight vectors $\bq$ to a given set of active experts $E$; then, the convex combination
used by $\cS_\eta$ at round $t$ can be written as $\bp_t = \bw_t^{E_t}$.

By convexity of the loss functions $\ell_t$, the regret with respect to some expert $j$ can be bounded as
\[
R_T(\cS_\eta,j) \leq \sum_{t=1}^T \left( \sum_{i \in E_t} w_{i,t}^{E_t} \, \ell_t(\delta_i) - \ell_t(\delta_j) \right) \ind_{ \{ j \in E_t \} }\,.
\]
Hoeffding's lemma (see, e.g., \cite[Lemma~A.1]{CeLu06}) entails that for all $t$ such that $j \in E_t$,
\begin{eqnarray*}
\sum_{i \in E_t} w_{i,t}^{E_t} \, \ell_t(\delta_i) & \leq & - \frac{1}{\eta} \ln \left( \sum_{i \in E_t} w_{i,t}^{E_t} \, e^{-\eta \ell_t(\delta_i)} \right)
+ \frac{\eta}{8} L^2 \\
& = & - \frac{1}{\eta} \ln \frac{w_{j,t} \, e^{-\eta \ell_t(\delta_j)}}{w_{j,t+1}} + \frac{\eta}{8} L^2
= \ell_t(\delta_j) - \frac{1}{\eta} \ln \frac{w_{j,t}}{w_{j,t+1}} + \frac{\eta}{8} L^2\,, \\
\end{eqnarray*}
where we used that the update of the weight
of an expert $j \in E_t$ can be rewritten by definition as
\[
w_{j,t+1} = w_{j,t} \, e^{- \eta \ell_t(\delta_j)} \frac{1}{\sum_{k \in E_t}
w_{k,t}^{E_t} \, e^{- \eta \ell_t(\delta_k)}}\,.
\]
For $j \not\in E_t$, we have that $w_{j,t+1} = w_{j,t}$, again by definition of the rule. Thus a telescoping sum appears and we get
\[
\sum_{t=1}^T \left( \sum_{i \in E_t} w_{i,t}^{E_t} \, \ell_t(\delta_i) - \ell_t(\delta_j) \right) \ind_{ \{ j \in E_t \} }
\leq - \frac{1}{\eta} \ln \frac{w_{j,1}}{w_{j,T+1}} + \frac{\eta}{8} L^2 \, \sum_{t=1}^T \ind_{ \{ j \in E_t \} }\,.
\]
The proof is concluded by noting that $w_{j,1} / w_{j,T+1} \geq 1/N$ as $w_{j,1} = 1/N$ and $w_{j,T+1} \leq 1$.
\end{proof}

\section{Proof of Theorem~\ref{prop:FSEWA}}
\label{sec:app}

The following proof is a straightforward adaptation of the techniques presented in~\cite[Section~5.2]{CeLu06}.
Its only merit is to show how the share update was obtained in Figure~\ref{fig:FSrule}.

\begin{proof}
We first note that by convexity of the $\ell_t$,
\begin{equation}
\label{eq:FSproof1}
\max_{j_1^T \in \cL_m} {R}_{T} \bigl( \cF_{\eta,\alpha}, j_1^T \bigr)
\leq \sum_{t=1}^T \left( \sum_{i \in E_t} p_{i,t} \ell_t(\delta_i) - \ell_t\bigl(\delta_{j_t}\bigr) \right).
\end{equation}
We now use the same proof scheme as in \cite[Section~5.2]{CeLu06} and show that the rule $\cF_{\eta,\alpha}$
is simply an efficient implementation of the rule that would, at each round $t$, choose a convex weight
vector $\bp'_t$ with components proportional to
\begin{numcases}{p'_{j,t} \,\, \propto \,\, w'_{j,t-1} \, = }
\nonumber
0 & if $j \not\in E_t$, \\
\nonumber
\sum_{j_1^T \in \cL} \nu \bigl( j_1^T \bigr) \, e^{- \eta \sum_{s=1}^{t-1} \ell_s(j_s)} \, \ind_{ \{ j_t = j \} }
& if $j \in E_t$,
\end{numcases}
where $\nu$ is some prior probability distribution over $\cL$, to be defined below.
It then follows from \cite[Lemma~5.1]{CeLu06} that for all $j_1^T \in \cL$,
\begin{equation}
\label{eq:FSproof2}
\sum_{t=1}^T \left( \sum_{i \in E_t} p'_{i,t} \ell_t(\delta_i) - \ell_t\bigl(\delta_{j_t}\bigr) \right)
\leq \frac{1}{\eta} \ln \frac{1}{\nu \bigl( j_1^T \bigr)} + \frac{\eta L^2 T}{8}\,.
\end{equation}
To get the stated bound, we thus need, one the one hand, to define the distribution $\nu$,
and on the other hand, to show that $\cF_{\eta,\alpha}$ indeed performs the efficient implementation indicated above.

[First part: \emph{Definition of $\nu$}]~~In
the sequel we denote by $|E|$ the cardinality of a subset $E$ of $\{ 1,\ldots,N \}$.
We fix a real number $\alpha \in [0,1]$ and
consider the following probability distribution $\nu$ over the sequences of (legal and illegal) experts, i.e., over $\{1,\ldots,N\}^T$.
For each element $j_1^T \in \cL$, we denote by $m$ its size, by $t_1,\ldots,t_m$
the instances $1 \leq t \leq T-1$ such that $j_t \ne j_{t+1}$, and by $\cT$ the set of instances
$1 \leq t \leq T-1$ such that $j_t = j_{t+1}$; we then set
\[
\nu \bigl( j_1^T \bigr) =
\frac{1}{|E_1|} \, \prod_{t \in \cT} \left(1-\alpha+\frac{\alpha}{|E_{t+1}|} \right)
\prod_{s=1}^m \left( \frac{\alpha}{|E_{t_s+1}|} \ind_{ \{ j_{t_s} \in E_{t_s+1} \} }
+ \frac{1}{|E_{t_s+1}|} \ind_{ \{ j_{t_s} \not\in E_{t_s+1} \} } \right);
\]
for $j_1^T \not\in \cL$, we set $\nu \bigl( j_1^T \bigr) = 0$.
This application $\nu$ indeed defines a probability distribution as can be seen by
introducing the uniform distribution $\mu_1$ over $E_1$ and the
following transition functions $\Tr_t : \{ 1,\ldots,N \}^2 \to [0,1]$;
for all $i,j$,
\begin{numcases}{\Tr_t(i \to j) =}
\label{eq:interpr1}
0 & if $j \not\in E_{t+1}$; \\
\label{eq:interpr2}
(1-\alpha) + \alpha\big/|E_{t+1}| & $j \in E_{t+1}$ and $i=j$; \\
\label{eq:interpr3}
\alpha\big/|E_{t+1}| & $j \in E_{t+1}$, $i \in E_{t+1}$, and $i \ne j$; \\
\label{eq:interpr4}
1\big/|E_{t+1}| & $j \in E_{t+1}$ and $i \not\in E_{t+1}$.
\end{numcases}
Its interpretation is as follows. We never switch to an inactive expert, as is ensured
by~(\ref{eq:interpr1}). If we can stay on the same expert (if the current expert remains active),
then we do so with a probability slightly larger than $1-\alpha$, see~(\ref{eq:interpr2}).
If we could have stayed on the same expert, then (\ref{eq:interpr1}) indicates that
we switch with probability $\alpha/|E_{t+1}|$ to a different expert in $E_{t+1}$.
Finally, (\ref{eq:interpr4}) controls the case when the current expert becomes inactive
and we need to switch to a new expert for the compound expert to be legal.

Now, we note that for all $i$ and $t$, by distinguishing whether $i \in E_{t+1}$ or $i \not\in E_{t+1}$,
\[
\sum_{j=1}^N \Tr_t(i \to j) = 1
\]
and that, for all $j_1^T \in \{1,\ldots,N\}^T$ (all of them--the legal and the illegal ones),
\begin{equation}
\label{eq:markov}
\nu \bigl( j_1^T \bigr) = \mu_1(j_1) \, \prod_{t=1}^{T-1} \Tr_t\bigl( j_t \to j_{t+1} \bigr)\,.
\end{equation}

To prove the stated bound, assuming we have proven as well that $\bp_t = \bp'_t$ for all $t$ (which we do below,
in the second part of the proof), it suffices to combine (\ref{eq:FSproof1}) and~(\ref{eq:FSproof2})
with the following immediate lower bound on the $\nu \bigl( j_1^T \bigr)$,
\[
\nu \bigl( j_1^T \bigr) \geq
\frac{1}{N} \left( \prod_{t \in \cT} (1-\alpha) \right)
\left( \prod_{s=1}^m \frac{\alpha}{N} \right) = \frac{1}{N} \, (1-\alpha)^{T-m-1} \left( \frac{\alpha}{N} \right)^{m},
\]
which we obtained by upper bounding all cardinalities $|E_t|$ by $N$ in the
definition of $\nu$ and by using $0 \leq \alpha \leq 1$. (The obtained bound
is actually exactly the one of \cite[Theorem~5.2]{CeLu06}, due to the loose way we lower bounded $\nu$.)

[Second part: \emph{Proof of the efficient implementation}]~~The
proof goes by induction and mimics exactly the one of~\cite[Theorem~5.1]{CeLu06}.
It suffices to show that for all $j \in \{ 1,\ldots,N \}$ and $t \in \{ 0, \ldots, T-1 \}$,
one has $w_{j,t} = w'_{j,t}$.
To do so, we first note that thanks to~(\ref{eq:markov}), the distribution $\nu$
can be interpreted as the distribution of an inhomogeneous Markov process, hence (\ref{eq:markov})
indicates the distribution that $\nu$ induces over $\{ 1,\ldots,N \}^s$, for all $1 \leq s \leq T$;
the latter is given by simply replacing $T$ by $s$ in~(\ref{eq:markov}).
We can therefore rewrite $w'_{j,t}$ as
\begin{equation}
\label{eq:defwprime}
w'_{j,t} = \sum_{j_1,\ldots,j_{t+1}} \nu \bigl( j_1^{t+1} \bigr) \, e^{- \eta \sum_{s=1}^{t} \ell_s(j_s)} \, \ind_{ \{ j_{t+1} = j \} }\,,
\end{equation}
where the first sum is (indifferently) taken over $\{ 1,\ldots,N \}^{t+1}$
or $E_1 \times \ldots \times E_{t+1}$. For $t = 0$, we get
\[
w'_{j,0} = \sum_{j_1=1}^N \nu(j_1) \, \ind_{ \{ j_1 = j \} } = \mu_1(j) = w_{j,0}\,,
\]
by definition of $\nu$ and of the $w_{j,0}$ (we recall that $\mu_1$ denotes the uniform distribution over $E_1$).
Now, we assume that for some $t \geq 1$, we have proved that $w_{i,t-1} = w'_{i,t-1}$ for all
$i \in \{ 1,\ldots,N \}$. For $j \in E_{t+1}$, by the share update in Figure~\ref{fig:FSrule} and by the induction hypothesis,
\begin{align*}
w_{j,t} = \ & \frac{1}{\abs[E_{t+1}]}\sum_{i \in E_t\setminus E_{t+1}}\!\!\!w'_{i,t-1}\,e^{-\eta \ell_t(\delta_i)}
+ \frac{\alpha}{\abs[E_{t+1}]}\sum_{i \in E_t\cap E_{t+1}}\!\!\!w'_{i,t-1}\,e^{-\eta \ell_t(\delta_i)} \\
& \ \ + (1-\alpha)\,\ind_{ \{ j\in E_t\cap E_{t+1} \} }\, w'_{j,t-1}\,e^{-\eta \ell_t(\delta_j)}\,.
\end{align*}
By definition of the transition functions~(\ref{eq:interpr1})--(\ref{eq:interpr4}),
this equality can be rewritten as
\[
w_{j,t} = \sum_{i \in E_t} w'_{i,t-1}\,e^{-\eta \ell_t(\delta_i)} \, \Tr_t(i \to j)\,.
\]
Substituting~(\ref{eq:defwprime}) in this equality, we get
\begin{eqnarray*}
w_{j,t} & = & \sum_{j_1,\ldots,j_t} \sum_{i \in E_t} \nu \bigl( j_1^{t} \bigr) \ind_{ \{ j_{t} = i \} } \Tr_t(i \to j)
\, e^{- \eta \sum_{s=1}^{t-1} \ell_s(j_s)} e^{-\eta \ell_t(\delta_i)} \\
& = & \sum_{j_1,\ldots,j_t} \nu \bigl( j_1^{t} \bigr) \Tr_t(j_t \to j) \, e^{- \eta \sum_{s=1}^{t} \ell_s(j_s)} \\
& = & \sum_{j_1,\ldots,j_t,j_{t+1}} \nu \bigl( j_1^{t+1} \bigr) \, \ind_{ \{ j_{t+1} = j \} } \, e^{- \eta \sum_{s=1}^{t} \ell_s(j_s)}
\,\, = \,\, w'_{j,t}\,,
\end{eqnarray*}
where the last but one equality follows from~(\ref{eq:markov}).
For $j \not\in E_{t+1}$, by definitions, $w_{j,t} = 0$ and $w'_{j,t} = 0$. This concludes this proof.
\end{proof}

\section{Proof of Corollary~\ref{cor:FSEG}}
\label{sec:app3}

This proof uses the same methodology as the one of Corollary~\ref{cor:ewa-spec}.

\begin{proof}
We fix a compound weight vector $\bq_1^T \in \cC_m$ and denote
by $\cL \bigl( \bq_1^T \bigr) \subseteq \cL_m$ the set of compound experts $j_1^T$ that are compatible
with $\bq_1^T$ in the following sense: denoting by $t_1,\ldots,t_m$ the time instances
$1 \leq s \leq T-1$ such that $\bq_{s} \ne \bq_{s+1}$, the elements $j_1^T$ in $\cL \bigl( \bq_1^T \bigr)$ are
characterized by the fact that $j_{s} \ne j_{s+1}$ only if $s = t_k$ for some $k \in \{ 1,\ldots, m \}$.
We insist on the fact that this is a ``only if'' statement and not an ``if and only if'' statement;
this means that the switches in the sequences $j_1^T \in \cL \bigl( \bq_1^T \bigr)$
can only occur (but are not bound to occur) at the indexes of the switches in $\bq_1^T$.

Now, we recall that by the gradient trick recalled in Section~\ref{sec:subgrad},
\[
R_{T} \bigl( \cF^{\grad}_{\eta,\alpha}, \bq_1^T \bigr)
\leq \tR_{T} \bigl( \cF^{\grad}_{\eta,\alpha}, \bq_1^T \bigr) =
\sum_{t=1}^T \Bigl( \tl_t(\bp_t) - \tl_t(\bq_t) \Bigr) \,.
\]
Since the $\tl_t$ are linear over $\cX$, the last expression can be upper bounded by
\[
\sum_{t=1}^T \Bigl( \tl_t(\bp_t) - \tl_t(\bq_t) \Bigr)
\leq \max_{j_1^T \in \cL \bigl( \bq_1^T \bigr)} \sum_{t=1}^T \Bigl( \tl_t(\bp_t) - \tl_t \bigl( \delta_{j_t} \bigr) \Bigr)\,,
\]
which shows that in particular,
\[
\sum_{t=1}^T \Bigl( \tl_t(\bp_t) - \tl_t(\bq_t) \Bigr)
\leq \max_{j_1^T \in \cL_m} \sum_{t=1}^T \Bigl( \tl_t(\bp_t) - \tl_t \bigl( \delta_{j_t} \bigr) \Bigr)
= \max_{j_1^T \in \cL_m} \tR_{T} \bigl( \cF^{\grad}_{\eta,\alpha}, j_1^T \bigr)\,.
\]
The proof is concluded by noting that Theorem~\ref{prop:FSEWA} exactly ensures that the rule $\cF^{\grad}_{\eta,\alpha}$
is such that
\[
\max_{j_1^T \in \cL_m} \tR_{T} \bigl( \cF^{\grad}_{\eta,\alpha}, j_1^T \bigr)
\leq \frac{m+1}{\eta} \ln N + \frac{1}{\eta}
\ln \frac{1}{\alpha^m \, (1-\alpha)^{T-m-1}} + \frac{\eta}{8} (2G)^2 T\,.
\]
\end{proof}

\section{Sketch of a regret bound on the operational adaptation $\cW_\eta$ of $\cE_\eta$}
\label{sec:operadapt}

We provide a proof by approximation and show that the regret of
$\cW_\eta$ is bounded by the regret of $\cE_\eta$ plus some small term.
To do so, we compare the definitions~(\ref{eq:defEWA})
and~(\ref{eq:defEWAext}), e.g., in the case when $p_{j,0} = 1/24$ for all experts $j$.

Since $R_{48 \lfloor (t-1)/48 \rfloor}( \cE_\eta, j)$ and
$R_{t-1}( \cE_\eta, j)$ differ by at most $47$ instantaneous regrets, each of
which is bounded between $-B^2$ and $B^2$, the ratio between the numerators
of~(\ref{eq:defEWA}) and~(\ref{eq:defEWAext}), as well as the one between
their denominators, lie in the interval
$\bigl[ e^{- 47 \eta B^2}, \,\, e^{47 \eta B^2} \bigr]$. Therefore, the ratios of the weights
defined in~(\ref{eq:defEWA}) and~(\ref{eq:defEWAext}) are in the interval
$\bigl[ e^{-94 \eta B^2}, \,\, e^{94 \eta B^2} \bigr]$.
Thus, using a gradient bound, the difference between the regrets of interest can be bounded as
\[
R_T( \cW_\eta, j) - R_T( \cE_\eta, j) \leq 2 B^2 \max \Bigl\{ e^{\eta 94 B^2}-1, \,\, 1-e^{-\eta 94 B^2} \Bigr\} \, T\,,
\]
which, for $\eta$ small enough, is of the order of $B^4 \eta T$. Taking $\eta$ of the the order of $1/\sqrt{T}$,
which is also the optimal order of magnitude for the bound on $R_T( \cE_\eta, j)$
stated in Theorem~\ref{prop:ewa}, entails that $R_T( \cW_\eta, j) = O\bigl(\sqrt{T}\bigr) = o(T)$,
as asserted above.

\end{document}